\documentclass[journal]{IEEEtran}
\usepackage{amsmath,amsfonts}
\usepackage{algorithmic}
\usepackage{algorithm}
\usepackage{array}
\usepackage{textcomp}
\usepackage{stfloats}
\usepackage{url}
\usepackage{verbatim}
\usepackage{graphicx}
\usepackage{cite}
\usepackage{color}
\usepackage{amsthm, amsfonts}
\usepackage{algorithmic}
\usepackage{algorithm}
\usepackage{subfigure}
\usepackage{siunitx}
\usepackage{multirow}
\usepackage{boldline}

\usepackage{booktabs}

\usepackage{xcolor}

\begin{document}
\title{Robot-Assisted Deep Venous Thrombosis Ultrasound Examination using Virtual Fixture}
\author{
	\vskip 1em
	
	Dianye Huang, 
	Chenguang Yang, 
	Mingchuan Zhou,
	Angelos Karlas, 
	Nassir Navab, 
	Zhongliang Jiang

	\thanks{

		\textit{Corresponding author: Zhongliang Jiang}
		
		Dianye Huang, Zhongliang Jiang, and Nassir Navab are with the Chair for Computer Aided Medical Procedures and Augmented Reality (CAMP), Technical University of Munich (TUM), Garching, Germany (e-mail: dianye.huang@tum.de; zl.jiang@tum.de; nassir.navab@tum.de). 
		
		Chenguang Yang is with Bristol Robotics Laboratory, University of the West of England, Bristol, BS16 1QY, UK (e-mail: cyang@ieee.org). 
		
		Mingchuan Zhou is with the College of Biosystems of Engineering and Food Science, Zhejiang University, China (e-mail: mingchuan.zhou@in.tum.de). 
		
		Angelos Karlas is with the Institute of Biological and Medical Imaging, Helmholtz Zentrum Munchen, Neuherberg, Germany, and also the Department for Vascular and Endovascular Surgery, rechts der Isar University Hospital, Technical University of Munich, Germany (e-mail: angelos.karlas@tum.de).
	}
}

\maketitle

\begin{abstract}
Deep Venous Thrombosis (DVT) is a common vascular disease with blood clots inside deep veins, which may block blood flow or even cause a life-threatening pulmonary embolism. A typical exam for DVT using ultrasound (US) imaging is by pressing the target vein until its lumen is fully compressed. However, the compression exam is highly operator-dependent. To alleviate intra- and inter-variations, we present a robotic US system with a novel hybrid force motion control scheme ensuring position and force tracking accuracy, and soft landing of the probe onto the target surface. In addition, a path-based virtual fixture is proposed to realize easy human-robot interaction for repeat compression operation at the lesion location. To ensure the biometric measurements obtained in different examinations are comparable, the 6D scanning path is determined in a coarse-to-fine manner using both an external RGBD camera and US images. The RGBD camera is first used to extract a rough scanning path on the object. Then, the segmented vascular lumen from US images are used to optimize the scanning path to ensure the visibility of the target object. To generate a continuous scan path for developing virtual fixtures, an arc-length based path fitting model considering both position and orientation is proposed. Finally, the whole system is evaluated on a human-like arm phantom with an uneven surface. The code\footnote{Code: \url{https://github.com/dianyeHuang/RobDVTUS}} and intuitive demonstration video\footnote{Video: \url{https://www.youtube.com/watch?v=3xFyqU1rV8c}}  can be publicly accessed. 
\end{abstract}

\def\abstractname{Note to Practitioners}
\begin{abstract}
Robotic ultrasound (US) systems have attracted attention for various applications in the past decades. However, the existing studies are not mature and intelligent enough for some challenging applications, such as DVT exam, which requires rich contact interaction between patients and clinicians. To tackle with this challenge, this study presents a novel human-centric robotic DVT exam program using the technique of virtual fixture. The coarse-to-fine path planning module ensures the repeatability of US acquisitions carried out at different times. During DVT exam, the proposed continuous 6D path virtual fixture can guide clinicians to freely move the probe along the scan path while limiting the probe motion in other directions. In order to perform the compress-release exam, a decoupled position/force controller is developed to precisely generate the contact force conveyed by clinicians and to restrict the probe motion along the probe centerline. We believe such a robot-assisted system is a promising solution to take both advantages of robots about the accuracy and repeatability and human operators about the advanced physiological knowledge. 
\end{abstract}

\begin{IEEEkeywords}
deep venous thrombosis (DVT), robotic ultrasound, hybrid force motion control, virtual fixture, path planning
\end{IEEEkeywords}



\section{Introduction}
\begin{figure}[t]
     \centering
     \includegraphics[width=0.38\textwidth, angle=0]{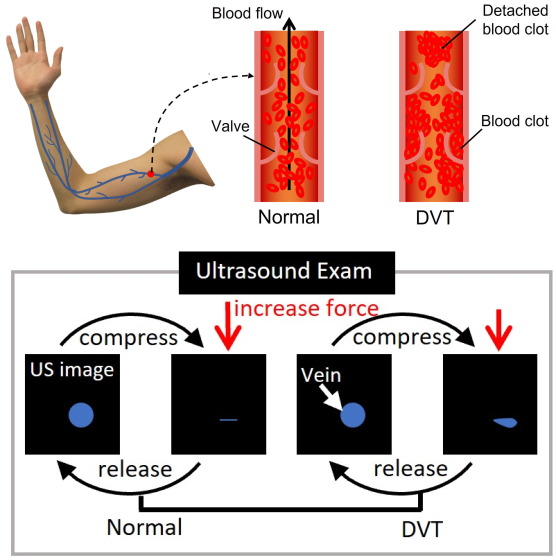}
     \vspace{-1em}
     \caption{An illustration of the DVT disease and the compression-release cycle of the DVT US exam.}
     \label{fig:DVT}
     \vspace{-1em}
\end{figure}

\IEEEPARstart{D}{eep} Venous Thrombosis (DVT) disease is one of the leading causes of postoperative morbidity and mortality~\cite{kakkar1990prevention}. It is an obstructive disease of the veins where the intravenous formation of blood clots may partially or completely block the blood return to the heart (see Fig. \ref{fig:DVT}) resulting in the life-threatening pulmonary embolism~\cite{kassai2004systematic}. DVT might be symptomatic (e.g., with leg swelling or pain) or completely asymptomatic, rendering its diagnosis a clinical challenge. 

The gold-standard for diagnosing DVT is ultrasound (US) imaging, for instance, the repetitive compression of the vein at different cross-sectional positions along all the veins \cite{kassai2004systematic}. Even if the recording of other US-based information, such as the intravenous color and spectral Doppler signal is also recommended nowadays, the compression test readout is the most useful information for the examiner. More specifically, failure to completely compress the venous lumen until it collapses indicates a vein thrombosis \cite{Salcudean2003}.

Although US imaging owns the advantages of being realtime, non-invasive and ionizing radiation-free, it does not come without limitations. The compression test is currently highly operator-dependent, which limits its diagnostic accuracy and sensitivity, especially in challenging and follow-up or re-test examinations aiming usually at quantifying the outcome of an applied therapy (e.g., anticoagulation). Such a follow-up control would require the conduction of the compression test in a maximally standardized way, ideally by the same operator in order to ensure directly and quantifiable comparability of the actual findings with previous ones. Thus, the development of approaches to limit the exertion of pressure to the minimum needed in order to set the diagnosis would be useful.

To improve imaging quality and repeatability, robotic technologies have been employed by accurately controlling the US acquisition parameters (probe orientation, contact point, and force)~\cite{jiang2020automatic, gilbertson2015force, jiang2020automatic_tie}. Recently, various robotic US systems (RUSS) have been developed for a wide range of clinical applications, like autonomous screening of breast~\cite{welleweerd2021out}, lung~\cite{ma2021autonomous}, and blood vessels~\cite{jiang2021autonomous, huang2023motion,jiang2023dopus}, etc.. These systems focused on automatizing the screening by using the US imaging feedback or/and external RGBD cameras. To autonomously navigate probe to standard planes, Jiang~\emph{et al.} presented a reward learning framework based on a few expert demonstrations~\cite{jiang2023intelligent}. A comprehensive summary can be found in a recent article detailing the fundamental and emerging technologies involved in developing RUSS~\cite{jiang2023robotic}.

\par

\begin{figure*}[t]
     \centering
     \includegraphics[width=0.98\textwidth, angle=0]{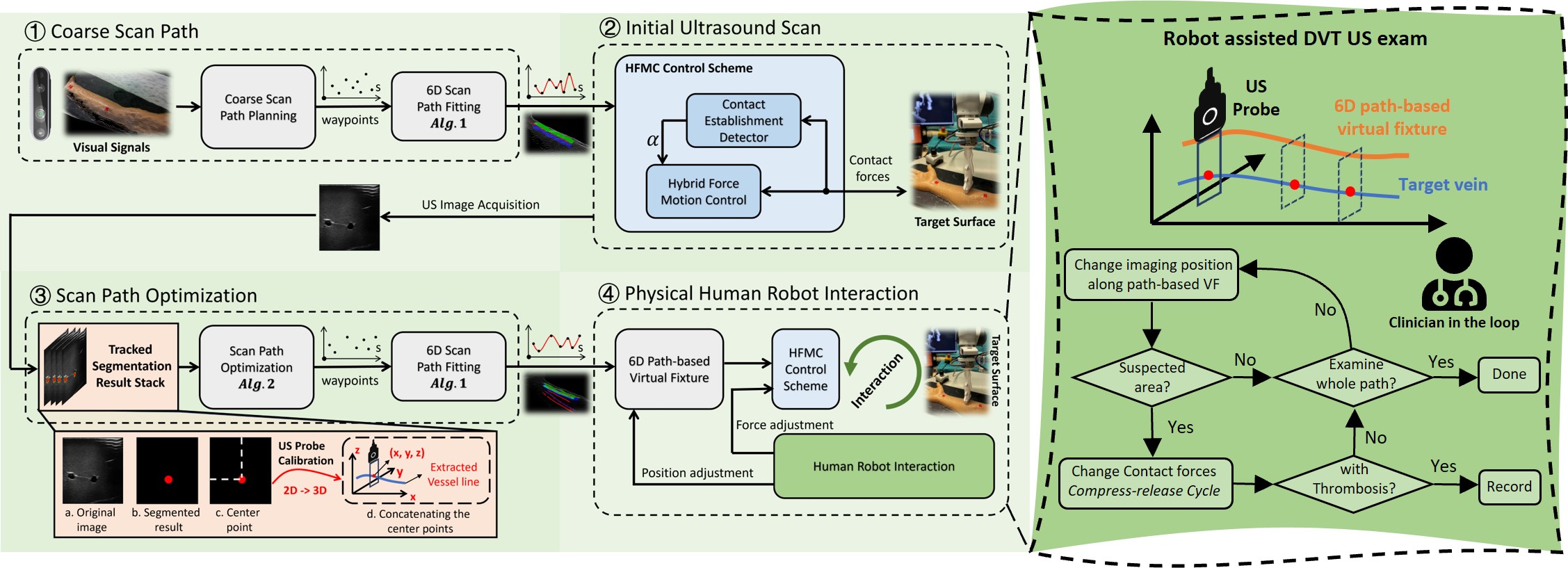}
     \caption{Workflow of the robot-assisted DVT exam. Left: algorithm blocks for deriving a 6D path-based virtual fixture. Right: an illustrative guide on performing the DVT US exam using the proposed system.}
     \label{fig:overall_workflow}
     \vspace{-1em}
\end{figure*}

However, due to the characteristics that DVT compression exams require rich interaction with the object vessel, the robot-assisted DVT examination has not been well-researched yet. To autonomously screen DVT, Meng~\emph{et al.} employed an RGBD camera to compute the scanning path, while the diagnosis was purely based on the segmentation of cross-sectional US images~\cite{Meng2015CYBER}. The missing performing compression exams will decrease the accuracy of diagnosis results. On the contrary, Guerrero~\emph{et al.} presented a RUSS with autonomous lumen contour detection and an external force sensor to capture the dynamical characters during the compression, while the scanning locations were manually determined~\cite{Salcudean2003}. 

\par
This work proposed a semi-autonomous DVT exam system to assist clinicians in quantifying the level of thrombus through a novel physical human-robot interaction (pHRI) interface based on the virtual fixture (VF) technique. The scanning path is autonomously determined coarse-to-fine using an RGBD camera and US images sequentially for coarse path planning and fine optimization. In addition, to guarantee the robot can accurately follow the scanning path and desired force, a novel hybrid force motion control scheme (HFMC) is developed in which the soft landing of the US probe onto the contact surface is taken into account. Furthermore, an arc-length scan path fitting method is proposed to generate a continuous 6D scanning trajectory with respect to a single variable. This enables the further development of a 6D continuous VF, which can assist sonographers to better perform compression exam for the target vein.  To the best of our knowledge, this is the first RUSS targeting DVT diagnosis with an practical pHRI to assist clinicians' manipulation of the US probe in desired direction (both scanning and compressing directions). 

\section{Related Work}

\subsection{Hybrid Force/Motion Control}
\par
Due to the safety concern for RUSS, impedance control has been widely used to realize compliant interaction between the probe and subjects~\cite{gilbertson2015force, jiang2021autonomous}. However, to accurately and simultaneously control the contact force and pose of the probe, impedance control requires accurate estimations of the environment's stiffness and damping information. To characterize the varying interaction environment, Pappalardo~\emph{et al.} used a simplified model to compute the interaction force $f = k_e (x-x_d)$ by considering elastic term ($k_e$ is environment stiffness)~\cite{interactionTRO2022}. The experimental results demonstrated that the exact value of $k_e$ is less important. The controller can still perform well when distinct stiffness parameters are selected. 

\par
Regarding the robot-assisted US examinations, the probe is often orientated by regulating the contact force along the probe axis and motion for the other five degree of freedom in Cartesian space. Thereby, the force and motion control laws can be designed separately, and assembled using projection matrices. To suppress force overshoot at the beginning of contact establishment, Halt \textit{et al.}~\cite{CASE2019ForceErr} proposed a controller based on prescribed performance control. This controller defined hard force error constraints and thanks to the \textit{log-type barrier Lyapunov function}~\cite{8643534}, the control input tends to be infinite to prevent constraint violation when the error approaches the predefined constraints. However, due to the numerical problem in computing and the hardware limitations in practices, the performance may decayed in real scenarios. To further address this challenge, we proposed a velocity-based force control law integrating the \textit{bounded barrier Lyapunov function}\cite{yang2021TNNLS} to avoid the numerical problem, and building a contact establishment detector to ensure a smooth transition from the free motion space to the contact space.

\subsection{Path Planning for US Scan}
\par
To autonomously plan a scanning path for RUSS, Huang \textit{et al.} presented a threshold-based method to extract the target of interest using an external camera~\cite{Huang2019TII}. The scanning path was computed based on the segmented surface on RGB images by considering the full coverage. Goel \textit{et al.} employed the Bayesian optimization method to identify the vessel-rich part of the leg to generate a scan path~\cite{9812410}. Jiang \textit{et al.} proposed a vision-based path planning pipeline that registers a CT template with an annotated blood vessel to the online point cloud streamed from an RGBD camera~\cite{jiang2022towards}. Then, a scanning path for the target vessel was determined by transferring the planned path from CT template to various volunteers. However, for the compression exam of DVT-US exam, these methods cannot guarantee that the US probe is placed above the blood vessel as the external camera cannot observe the internal patient-specific location of the target anatomical structure. Therefore, we proposed a coarse-to-fine pipeline for US scanning path planning method that sequentially utilize the external visual signals and the interior B-mode US images. 

\subsection{Physical Human-Robot Interaction}
To fully take advantage of the physiological knowledge (e.g., suitable contact force) from clinicians, a semi-autonomous RUSS is designed to allow physical interaction by operators. To assist the clinicians to qualitatively perform the DVT-US compression exams, virtual fixture (VF) is employed to restrict the robot's workspace, which usually consists of constraints definition, evaluation, and enforcement \cite{bowyer2013active}. To ensure safety and/or contact accuracy, VF has been widely used in the field of surgical robotics. Feng \emph{el al.} developed both guidance and forbidden VFs for high-precise polyp dissection~\cite{9605178}. Li \emph{et al.} constructed a protective VF from polygon mesh representations for the skull-cutting application~\cite{9341590}. 

\par
Regarding DVT-US examinations, the US probe should adhere to the scanning path to generate comparable results in various US compression exams. Thus, a 6D path virtual fixture is expected to ease the interaction. A scan path is often represented by a set of waypoints and orientation pairs. To obtain a continuous path, Tan~\emph{et al.} fitted the waypoints using the non-uniform rational B-spine (NURB) method~\cite{Tan2022TMRB}. However, this method is time-consuming and hard to conduct constraint evaluation for a path VF. Therefore, in our case, a spatially encoded curve fitting method is proposed, which reduces the 6D scan path to a 1D manifold space to correlate the examined cross-section with a one-dimensional variable, thus realizing a path VF. 

\subsection{Proposed Workflow}
The primary clinical indicators for the robot-assisted DVT-US exam system include the precision of motion and force control, and the system's ability to position the target vein at the central region of the acquired image. Precise control ensures force and posture alignment with clinician intentions during compression-release cycles. Furthermore, the system's optimization of the scanning path enhances the effectiveness and repeatability of DVT US exams. Consequently, this work focuses on the control design and the scan path determination.

The proposed robot-assisted DVT exam system's workflow depicted in the left side of Fig. \ref{fig:overall_workflow} consists of four phases. \emph{i).} A coarse discrete path is first determined using external visual signals captured from the RBG-D camera mounted at the robot's end-effector. By fitting the 6D discrete waypoints, an initial continuous scan path is derived. \emph{ii).} The next phase involves performing an US sweep scan. Simultaneously, a UNet model segments the acquired US image, enabling the locating of the target vein in the cross-sectional view. \emph{iii).} The initial scan path is then optimized using the segmentation results. After fitting the optimized discrete scan path, a 6D path-based virtual fixture is established. \emph{iv).} Finally, the sonographers can physically interact with the robot, adhering to the constraints imposed by the virtual fixture, to maneuver the US probe along the optimized scan path. They can also make adjustments to the contact forces in a smooth manner, drawing upon their expertise and experience.

\begin{figure}[t]
     \centering
     \includegraphics[width=0.35\textwidth, angle=0]{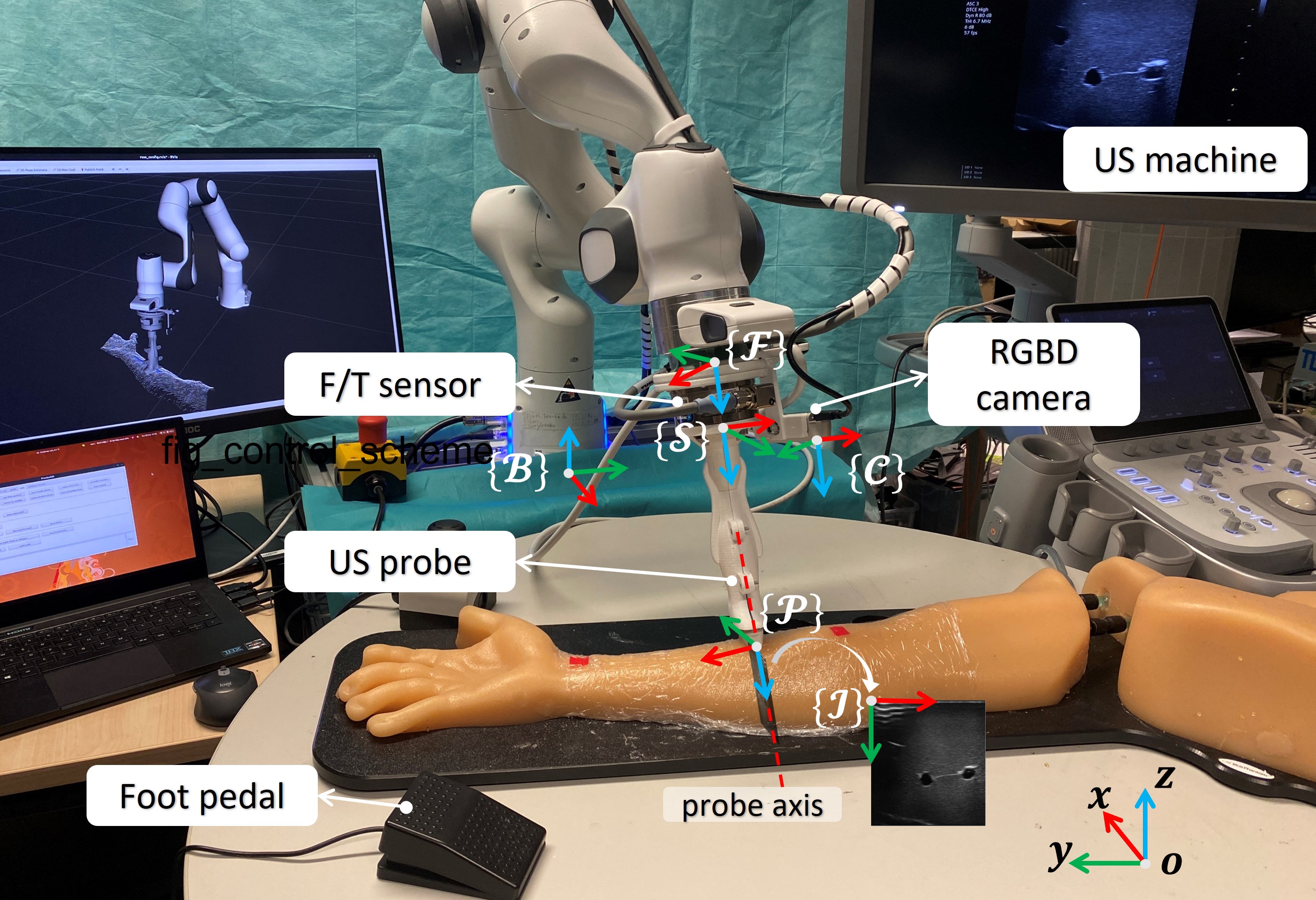}
     \caption{An illustration of system setup and coordinate frames.}
     \label{fig:system_setup}
     \vspace{-1em}
\end{figure}

\section{Force Motion Control Scheme}
\label{sec:HFMC}
Developing a highly accurate force controller is important in the context of the DVT exam. As depicted in Fig. \ref{fig:DVT}, clinicians perform a compression-release cycle by incrementally adjusting contact forces until one of the following conditions is met: i) the target vein fully closes, indicating the absence of thrombosis, or ii) the vein ceases to deform while remaining partially open. This cycle is repeated multiple times. The proposed framework aims to enhance the repeatability and utility of the DVT exam for long-term disease monitoring, which necessitates accurate force control. Therefore, this section elaborates on the design of the proposed hybrid force motion control scheme. The coordinate system is depicted in Fig. \ref{fig:system_setup}, including frames of robot's base $\{\mathcal{B}\}$, robot's Flange $\{\mathcal{F}\}$, force/torque (F/T) sensor $\{\mathcal{S}\}$, RGBD camera $\{\mathcal{C}\}$, US probe $\{\mathcal{P}\}$ and US image $\{\mathcal{I}\}$. Unless otherwise stated, vectors and matrices are described in $\{\mathcal{B}\}$, and the left upper script is omitted for brevity. Three different calibrations are needed to correlate the transformations among these frames, including: 

\begin{enumerate}
\item Eye-in-hand calibration \cite{hecalib1989Tsai}: calibrating ${^\mathcal{F}_\mathcal{C}}\mathbf{T}$ to transform the scan path expressed from $\{\mathcal{C}\}$ to $\{\mathcal{F}\}$.
\item F/T sensor calibration \cite{forcecomp2017}.: compensating for the payload gravity to derive $^\mathcal{S}\mathbf{f}_{cali}$ and calibrating $^\mathcal{P}_\mathcal{S}\mathbf{T}_{twist}$ to obtain the contact force $^\mathcal{P}\mathbf{f}_{cali}={^\mathcal{P}_\mathcal{S}\mathbf{T}_{twist}}{^\mathcal{S}\mathbf{f}_{cali}}$.
\item US probe calibration \cite{jiang2022precise}: calibrating ${^\mathcal{F}_\mathcal{P}}\mathbf{T}$ to transfer the 2D pixel location into 3D space by computing ${^\mathcal{B}_\mathcal{I}}\mathbf{T}={^\mathcal{B}_\mathcal{F}\mathbf{T}}~{^\mathcal{F}_\mathcal{P}\mathbf{T}}~{^\mathcal{P}_\mathcal{I}\mathbf{T}}$, where
\begin{equation*}
    {^\mathcal{P}_\mathcal{I}}\mathbf{T}=
    \left[
    \begin{array}{cccc}
         0 & 0 & -1 & 0 \\
         0 & L_p/W_{img} & 0 & -L_p/2 \\
         D_{img}/H_{img} & 0 & 0 & 0\\
         0 & 0 & 0 & 1
    \end{array}\right]
\end{equation*}
$L_p$, $D_{img}$, and $H_{img}\times W_{img}$ denote the footprint, imaging depth and imaging resolution of the US probe.
\end{enumerate}

\subsection{Hybrid Force Motion Control Law}
Since the force and the motion control axis are orthogonal, we first decoupled them by projecting the tracking errors into their respective space, and then assembled the designed two control laws. We assume that the nonlinear gravity, Centrifugal and Coriolis torques of the robot dynamic are fully resolved. Thus, the hybrid force motion control law alongside Eq. (\ref{eq:err_transform})$\sim$(\ref{eq:force_control_law}) is given below:
\begin{equation}
    \begin{aligned}
        \mathbf{\tau}&= 
        \mathbf{J}^T\bigg(-\mathbf{K}_p^c\mathbf{x}_e-\mathbf{K}_d^c\mathbf{\dot{x}}\bigg) +
        \mathbf{N}\bigg(\mathbf{K}_p^q\mathbf{q}_e-\mathbf{K}_d^q\dot{\mathbf{q}}\bigg) \\
        \mathbf{x}_e &= \bigg[\left(\mathbf{x} - \left(\mathbf{P}_m\mathbf{x}_d^\prime + \int{\mathbf{n}_z v_f}dt\right)\right)^T~~ \mathbf{e}_o^T\bigg]^T
    \end{aligned}
    \label{eq:hfmc_law}
\end{equation}
where $\tau$ is a torque-based control input, $\mathbf{J}$ is the Jacobian matrix of $\{\mathcal{P}\}$ w.r.t. $\{\mathcal{B}\}$ over the joint configuration $\mathbf{q}$; $\mathbf{N}=\mathbf{I}_{7\times 7}-\mathbf{J}^T(\mathbf{J}^T)^{-1}$ projects the joint torques that try maintaining the desired joint configuration $\mathbf{q}_d$ into the null space of the Cartesian control with $\mathbf{q}_e:=\mathbf{q}-\mathbf{q}_d$, and $\mathbf{K}_p^q$/$\mathbf{K}_d^q$ being the joint stiffness/damping configurations. For the Cartesian control, $\mathbf{K}_p^c$/$\mathbf{K}_d^c$, $\mathbf{x}$/$\textbf{x}_d^\prime$ and $\mathbf{q}$/$\mathbf{q}_d$ represent the Cartesian stiffness/damping settings, current/desired position and quaternion respectively; $\mathbf{x}_e$ denotes the pose error where the orientation error $\mathbf{e}_o := -{^{\mathcal{B}}_{\mathcal{P}}\mathbf{R}}~ vec(\overline{\mathbf{q}}*\mathbf{q}_d)$ is formulated by taking $vec(\cdot)$ the vector part of the transformed quaternion difference with ${^\mathcal{B}_\mathcal{P}\mathbf{R}}$ being the rotational matrix. ${^{\mathcal{B}}\mathbf{R}_p}$ can be expanded as $\big[~\mathbf{n}_x~\big|~\mathbf{n}_y~ \big|~\mathbf{n}_z~\big]$ where $\mathbf{n}_x, \mathbf{n}_y, \mathbf{n}_z \in \mathbb{R}^{3\times 1}$ and $\mathbf{n}_z$ represents the force control direction. Regarding the position error, $\mathbf{P}_m:=\mathbf{I}_{3\times 3}-\mathbf{n}_z \mathbf{n}_z^T$ filters the desired position such that an updated desired position adapting along $\mathbf{n}_z$ is formulated to regulate the contact force. In this paper, $\mathbf{K}_p^c=diag(1200\mathbf{I}_{3\times3}, 90\mathbf{I}_{3\times3})$, $\mathbf{K}_p^q=1e^{-3}\mathbf{I}_{7\times7}$, the damping ratio is selected to be $0.8$, $\mathbf{q}_d$ equals to the initial joint configuration before the sweep scanning. 

In Eq. (\ref{eq:hfmc_law}), the probe motion is regulated under the hierarchical Cartesian impedance control framework while the contact force is controlled by adapting the velocity $v_f$ (to be designed in \emph{Sec. \ref{sec:force_ctrl_law}}) along $\mathbf{n}_z$. 

\begin{figure}[!t]
    \centering
    \includegraphics[width=0.44\textwidth, angle=0]{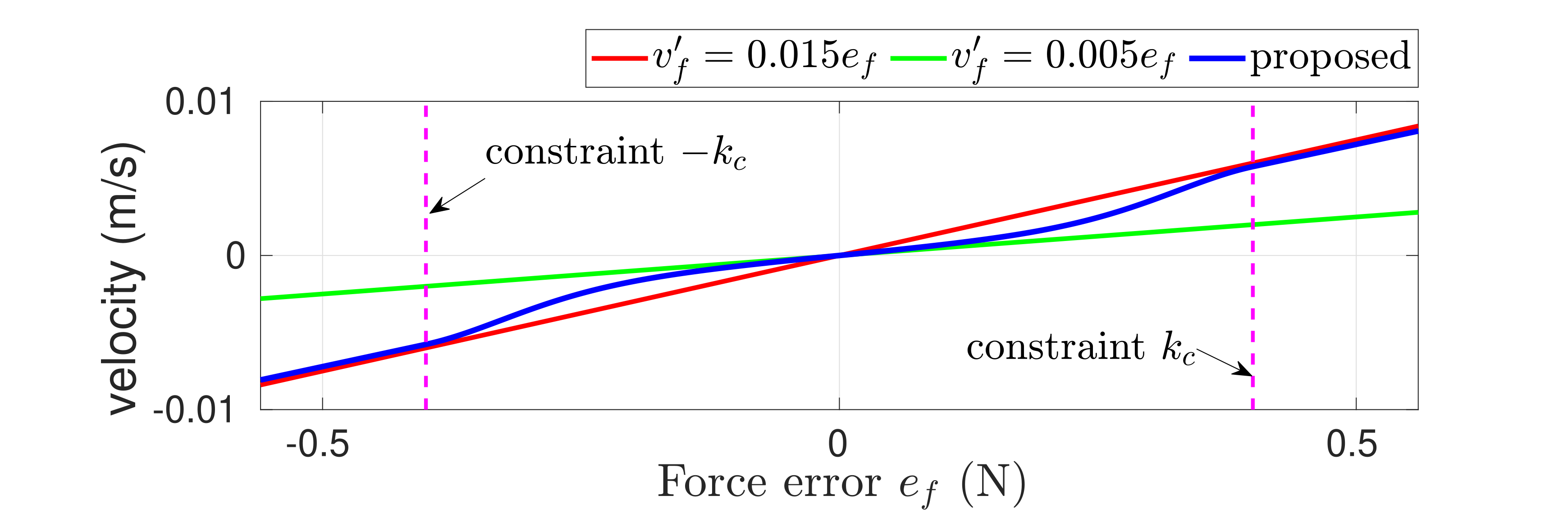}
    \vspace{-1em} 
    \caption{A plot of force control law in Eq. (\ref{eq:ctrl_law_force}) \textit{w.r.t.} $e_f$.}
    \label{fig:errtf}
    \vspace{-1em}
\end{figure}

\subsection{Velocity-based Force Control Law}
\label{sec:force_ctrl_law}
As in \cite{interactionTRO2022} and \cite{CASE2019ForceErr}, we adopted a spring model to approximate the interaction dynamics: $f = k_e(x-x_{env})$ where $x_{env}$ is the boundary of the undeformed tissue, $x$ is the current position of the robot's end-effector, $k_e$ is the environment stiffness term that reflects the level of resistance to the deformation. The actual contact force $f\in\mathbb{R}$ along the probe axis is measured by:
$f = \mathbf{n}_s^T~{^{\mathcal{B}}_{\mathcal{P}}\mathbf{T}_{tw}}{^{\mathcal{P}}\mathbf{f}_{c}}$
where $\mathbf{n}_s := \left[\mathbf{n}_z^T,~0,~0,~0\right]^T$ filters the measured torques, and the twist transformation ${^{\mathcal{B}}_{\mathcal{P}} \mathbf{T}_{tw}}\in\mathbb{R}^{6\times 6}$ transforms the calibrated contact force ${^\mathcal{P} \mathbf{f}_c}$  from  $\{\mathcal{P}\}$ to $\{\mathcal{B}\}$.

Inspired by \cite{CASE2019ForceErr} where the force error is transformed before being fed into the controller, in this paper, a \textit{bounded barrier Lyapunov function}~\cite{yang2021TNNLS} based error transformation is designed to overcome the potential numerical problem. The force tracking error $e_f$ is considered as a soft constrained variable $e_f = f-f_d$ subjected to $|e_f|<k_c$, where $k_c\in\mathbb{R}^{+}$ denotes the user-defined force error constraint, $f_d$ and $f$ denote the desired and actual contact forces, respectively. 

Error transformation is then designed as below:
\begin{equation}
    \varepsilon_f = |e_f| k_n \mathcal{T}(z) z
    \label{eq:err_transform}
\end{equation}
where $\mathcal{T}(z) = \frac{k_h k_s^2}{k_c}\frac{1-z^2}{(1-k_s^2z^2)^2} > 0$, $\mathcal{L}im(e_f, -k_c, k_c) := \max(\min(e_f, k_c), -k_c)$, $z = \tanh\left(k_h\mathcal{L}im\big(e_f, -k_c, k_c\big)/k_c\right)$ is an intermediate variable introduced to avoid the numerical problem of the error transformation; $k_n$ normalizes the transformed error; $k_s$ determines the derivative of $\varepsilon_f$ \textit{w.r.t.} $e_f$. Note that once $k_s$ is selected, $k_n$ and $k_h$ in Eq. (\ref{eq:err_transform}) remain constant and are computed according to the following equation: 
\begin{equation*}
    k_n = \frac{1}{\mathcal{T}(\zeta)\zeta},~ k_h = \ln\left(\sqrt{\frac{1+\zeta}{1-\zeta}}\right), ~\zeta  = \sqrt{\frac{3 k_s-\sqrt{4-3 k_s^2}}{2 k_s}} 
\end{equation*}

Below is the force control law in the contact space.
\begin{equation}
    v_f^\prime = -k_{mf}\varepsilon_f - k_f e_f
    \label{eq:ctrl_law_force}
\end{equation}
where $k_{mf}$ and $k_f$ are feedback gains for the transformed force error $\varepsilon_f$ and actual force error $e_f$ respectively; $v_f^\prime$ represents the velocity command for regulating the contact force in the contact space. Fig. \ref{fig:errtf} plots the control input of Eq. (\ref{eq:ctrl_law_force}) \textit{w.r.t} $e_f$ where parameters are selected empirically, $k_c=0.4$, $k_s = 0.99$ for the error transformation, and $k_{mf}=0.008$, $k_f=0.0065$ for the force control law in Eq. (\ref{eq:ctrl_law_force}). Note that the proposed control law smoothly transits between the nonlinear and linear feedback gains wherein the predefined constraints $\pm k_c$ serve as the transition boundaries. 

\subsection{Soft Landing Approach}
In order to guarantee soft landing of US probe onto the target surface, we designed and incorporated a low-pass filter into the control scheme (input: $f$, output: $\alpha$, see \textit{Contact Establishment Detector} in Fig. \ref{fig:overall_workflow}).
\begin{equation}
    \begin{aligned}
    \dot{\alpha} + f_{\bar{\alpha}}k_\alpha \alpha = k_\alpha \mathcal{C}lamp\left(f, ~f_{\underline{\alpha}}, ~f_{\bar{\alpha}}\right) \\
    \end{aligned}
    \label{eq:contact_impact}
\end{equation}
where 
\begin{equation*}
    \mathcal{C}lamp(f,~f_{\underline{\alpha}}, ~f_{\bar{\alpha}})=
\begin{cases}
    0& if~f<f_{\underline{\alpha}}\\
    f& if~f\in [f_{\underline{\alpha}},~f_{\bar{\alpha}}]\\
    f_{\bar{\alpha}} & if~f>f_{\bar{\alpha}}
\end{cases}
\end{equation*}
$\mathcal{C}lamp(\cdot, \cdot, \cdot)$ is introduced to clamp $f$ within the predefined limits $f_{\underline{\alpha}}$ and $f_{\bar{\alpha}}$, and the contact is sensed when $f > f_{\underline{\alpha}}$; $k_\alpha$ determines the sensitivity of the filter; $\alpha\in[0,~1]$ serves as a signal indicating the contact establishment. The final force control law along the probe axis is given by:
\begin{equation}
    v_f = \alpha v_f^\prime + (1-\alpha)v_0
    \label{eq:force_control_law}
\end{equation}
where $\alpha$ is emerged into $v_f$ to govern the control law transition between a predefined constant velocity $v_0$ for approaching the scan target in free motion space and the proposed control law $v_f^\prime$ in  Eq. (\ref{eq:ctrl_law_force})  for contact force regulation in the contact space. In this paper, $f_{\underline{\alpha}}=1.0$, $f_{\bar{\alpha}}=2.0$, $k_\alpha=10$, $v_0=0.015$.  

\begin{algorithm}[b]
	\renewcommand{\algorithmicrequire}{\textbf{Input:}}
	\renewcommand{\algorithmicensure}{ \textbf{Output:}}
	\caption{Scan Path Fitting}
	\label{alg:dmp_curve_fitting}
	\begin{algorithmic}[1]
	    \REQUIRE Scan point and normal vector pairs $(\mathcal{P},~\mathcal{N})$,\\
	            ~~~ Euclidean distance threshold $\delta_p$ ($1~mm$ by default).
	    \ENSURE Learned $\mathcal{F}_p(s)$, $\mathcal{F}_Q(s)$ and $s_N$.
	    \STATE $\mathcal{P}_{new}\gets\mathcal{P}$ \textit{Linear interpolation} if $\|\mathcal{P}^{(i+1)}-\mathcal{P}^{(i)}\|>\delta_p$
	    \STATE $\mathbf{s}$, $s_N$ $\gets$ Eq. (\ref{eq:calc_s}).
	    \STATE $\mathcal{F}_p(s)$ $\gets$ Eq. (\ref{eq:dmp_pos_model}) with $\mathbf{s},~\mathcal{P}_{new}$. 
	    \STATE $\mathcal{Q}\gets\big\{\mathbf{R}^{(i)} = \left[~\bar{\mathbf{n}}_x~\big| ~\bar{\mathbf{n}}_y~\big|~\mathcal{N}^{(i)}~\right],~\bar{\mathbf{n}}_x^\prime = \mathcal{F}_p(\mathbf{s}^{(i)}+\delta_s/S_N) - \mathcal{F}_p(\mathbf{s}^{(i)})$, $\bar{\mathbf{n}}_y = \mathcal{N}^{(i)}\times\bar{\mathbf{n}}_x^\prime$, $\bar{\mathbf{n}}_x=\bar{\mathbf{n}}_y\times\mathcal{N}^{(i)}\big\}_i^N$
	    \STATE $\mathcal{Q}_{new}\gets\mathcal{Q}$ \textit{Slerp interpolation} if $\|\mathcal{P}^{(i+1)}-\mathcal{P}^{(i)}\|>\delta_p$
	    \STATE $\mathcal{F}_Q(s)$ $\gets$ Eq. (\ref{eq:dmp_ort_model}) with $(\mathbf{s}, ~\mathcal{Q}_{new})$.
	\end{algorithmic}  
\end{algorithm}

\begin{algorithm}[t]
	\renewcommand{\algorithmicrequire}{\textbf{Input:}}
	\renewcommand{\algorithmicensure}{\textbf{Output:}}
	\renewcommand{\algorithmicrepeat}{\textbf{repeat}~~//~\textit{sweep scan}}
	\newcommand{\INDSTATE}[1][1]{\STATE\hspace{#1\algorithmicindent}}
	\caption{Scan Path Optimization}
	\label{alg:refined_scan_path_planning}
	\begin{algorithmic}[1]
	    \REQUIRE $\mathcal{F}^{init}_p(s)$, $\mathcal{F}^{init}_Q(s)$, $s_N^{init}$, scan velocity $v_{s}~(m/s)$.
	    \ENSURE  Learned $\mathcal{F}_p(s)$, $\mathcal{F}_Q(s)$ and $s_N$.
	    \STATE {Set $s=0$, $i=0$}
        \REPEAT 
        \STATE $\mathbf{x}_d\gets\mathcal{F}_p^{init}(s)$,  $\mathbf{Q}_d\gets\mathcal{F}_Q^{init}(s)$
        \STATE Eq. (\ref{eq:hfmc_law}) $\gets$ Track $\mathbf{x}_d$ and $\mathbf{Q}_d$.
        \STATE $\mathcal{I}_{seg}\gets$ Segment the US image by a UNet \cite{jiang2022towards}.
        \STATE $\mathcal{P}^{(i)}_{ves}\gets$ Extract center point of the blood vessel.
        \STATE Update $s \gets s + v_s/s_N^{init} \Delta t$;~~update $i\gets i+1$
        \UNTIL{$s \ge 1.0$}
        \STATE $\mathcal{P}$, $\mathcal{N}\gets$ Project $\mathcal{P}_{ves}$ to $\mathcal{P}_{arm}$ by Eq. (\ref{eq:vessel_project}).
        \STATE $\mathcal{F}_p(s)$, $\mathcal{F}_Q(s)$, $s_N\gets$ \textit{Alg. \ref{alg:dmp_curve_fitting}} with $(\mathcal{P},~\mathcal{N})$.
	\end{algorithmic}  
\end{algorithm}

\section{Physical Human Robot Interaction Interface}
\label{sec:scan_path_planning}
This section elaborates on the path planning and pHRI strategies. As shown in Fig. \ref{fig:overall_workflow}, the discrete planned path on the scanned surface right above the target vein is determined in a coarse-to-fine manner using the external visual signals and US images. The 6D discrete waypoints are then fitted and parameterized over a one-dimensional variable $s$. This enables the continuous output to serve as a motion generator for the initial sweep scan, which captures US images for path optimization. Additionally, it serves as a 6D path virtual fixture for physical human-robot interaction. By correlating the interaction forces with the change of desired contact force and $s$, clinicians can smoothly adjust the imaging positions and contact forces during the compression exams.

\subsection{Arc-length-based 6D Path Fitting}
\label{sec:path_fitting}
\par
The scan path extracted from an external camera is usually a series of discrete waypoints and normal vectors of the contact surface. These points and vectors are supposed to be converted into a 6D trajectory before being fed into the HFMC. Inspired by the \textit{arc length-dynamic movement primitive} (AL-DMP) method\cite{GASPAR2018225}, which separates the spatial and temporal components of motion, we proposed a curve fitting model using \textit{arc-length} to remove temporal information and encode the scan path spatially. Note that the proposed path fitting algorithm does not mere fit the given data, but also parameterized the 6D scan path over a 1D variable. \textit{Alg. \ref{alg:dmp_curve_fitting}} summarizes the steps. 

\subsubsection{Accumulated Arc-Length Variable}
Given a set of discrete points $\mathcal{P}\in\mathbb{R}^{N\times 3}$. The normalized accumulated \textit{arc-length} variable $\mathbf{s}\in\mathbb{R}^{N\times1}$ is defined as follows:
\begin{equation}
    \mathbf{s}=\{s^{(k)}\}, ~s^{(k)}=\sum_{i=1}^{k}\|\mathcal{P}^{(k)}-\mathcal{P}^{(k-1)}\|^2/s_{N}
    \label{eq:calc_s}
\end{equation}
where $s^{(0)}=0$ , $s_{N}=\sum_{i=1}^{N}\|\mathcal{P}^{(k)}-\mathcal{P}^{(k-1)}\|^2$ and the upper right script $(k)$ denotes the $k$-th element of the variable. 

\subsubsection{Curve Fitting Model}
Given a set of input-output pairs ($\mathbf{s}\in\mathbb{R}^{N \times 1},~\mathcal{X}\in\mathbb{R}^{N \times 1}$). The one-dimensional fitting model is designed as follows: 
\begin{equation}
    \mathcal{F}(s) = \bar{s}\mathcal{X}_0 + s\mathcal{X}_g + u(s, \mathbf{w}) 
    \label{eq:apdx_cf_model}
\end{equation}
where $u(s, \mathbf{w}) = \frac{\sum^N_{i=0}\psi_i(s)w_i\mathcal{M}(s)}{\sum^N_{i=0}\psi_i(s)}$, $s\in(0, ~1)$ is a monotonically increasing variable, $\bar{s} := 1-s$, $\mathcal{X}_0$ and $\mathcal{X}_g$ are the first and last sample of $\mathcal{X}$, $u(s,~\mathbf{w})$ is a nonlinear term consisting of weighted \textit{Truncated Gaussian Functions} (TGF)~\cite{7759554}. Since the AL-DMP based on \textit{arc-length} suffers from not converging to a unique attractor point, we induced $\mathcal{M}(s) = s\left(1 - \exp(-a\bar{s})\right)$ to overcome this issue by modulating $s$ such that $\mathcal{F}(s=0) = x_0$ and as $s\to1,~\mathcal{F}(s=1)\to x_g$. $\mathbf{w}$ denotes a set of weights $w_i$ for TGFs and is learned by the \textit{Local Weighted Regression} (LWR) algorithm \cite{atkeson1997locally}.

\subsubsection{Position Fitting}
The position model is designed below:
\begin{equation}
    \mathcal{F}_p(s) = \bar{s}\mathbf{x}_0 + s\mathbf{x}_g + \mathbf{u}(s, \omega_p)
    \label{eq:dmp_pos_model}
\end{equation}
where $\mathbf{x}_0$/$\mathbf{x}_g\in\mathbb{R}^{3\times1}$ denotes the initial/goal positions, and $\omega_p:=\{\mathbf{w}_p^x,~\mathbf{w}_p^y,~\mathbf{w}_p^z\}$ is learned axis-by-axis.

\subsubsection{Orientation Fitting} Unit quaternion is chosen for orientation representation. Considering the geometric constraint, the orientation fitting model $\mathcal{F}_Q(s)$ is formulated below:
\begin{equation}
    \begin{aligned}
        \mathcal{F}_{Q}(s) &= \overline{\exp\left(\mathbf{e}_Q/2\right)}*\mathbf{Q}_g  \\
        \mathbf{e}_Q &= (1-s)\mathbf{e}_{Q_0} + s \mathbf{e}_{Q_g} + \mathbf{u}(s, \omega_Q) 
    \end{aligned}
    \label{eq:dmp_ort_model}
\end{equation}
where $\omega_Q:=\{\mathbf{w}_Q^x,~\mathbf{w}_Q^y,~\mathbf{w}_Q^z\}$ is learned axis-by-axis; $\overline{\mathbf{Q}}$ denotes the conjugate of $\mathbf{Q}:=\{\eta,~\mathbf{\epsilon}\} := \eta + x\mathbf{i} + y\mathbf{j} + z\mathbf{k}\in\mathcal{S}^3$. $\mathbf{e}_Q=2\log(\mathbf{Q}_g*\overline{\mathbf{Q}})$ denotes the deviation of the goal quaternion. $\mathbf{e}_{Q_0}/\mathbf{e}_{Q_g}$ denotes the first/last sample of goal quaternion deviation. The logarithmic and exponential operators (refer to \cite{pmlr-v100-koutras20a} for more details) are presented below:
\begin{equation*}
    \begin{aligned}
        \mathbf{r} &= \log(\mathbf{Q})=\begin{cases}
    \arccos(\eta)\frac{\epsilon}{\|\epsilon\|} & \|\epsilon\| > 0\\
    [0, 0, 0]^T & otherwise
    \end{cases}\\
    \mathbf{Q} &=\exp(\mathbf{r}) = \begin{cases}
    \left[\cos(\|\mathbf{r}\|), ~\sin(\|\mathbf{r}\|)\frac{\mathbf{r}^T}{\|\mathbf{r}\|}\right]^T & \|\mathbf{r}\|>0 \\
    [1, 0, 0, 0]^T & otherwise
    \end{cases}
    \end{aligned}
\end{equation*}

\subsection{6D Path-based Virtual Fixture}
\label{sec:pathVF}
\subsubsection{Initial Coarse Scan Path}
As shown in the workflow of Fig. \ref{fig:overall_workflow} and the intermediate results presented in Fig. \ref{fig:path_plan_res}, the initial scan path is planned using only visual signals. The skeleton of the arm phantom $\mathcal{P}$ is extracted and filtered after segmenting the arm phantom in a depth image \cite{jiang2022towards, LEE1994462}. The starting and ending scan points are then determined by the two manually attached stickers. 

The initial scan path is finally obtained and fitted by \textit{Alg. \ref{alg:dmp_curve_fitting}}. In this work, $h=3.0$, $\theta=3.5$ for all \textit{TFGs}. The number of kernels for $\mathcal{F}_p(s)$  and $\mathcal{F}_Q(s)$ are empirically set to be $41$ and $81$ respectively, balancing the computational burden and fitting performance.

\subsubsection{Optimized Scan Path}
\textit{Alg. \ref{alg:refined_scan_path_planning}} summarizes the steps of path optimization. The initial scan path $\mathcal{F}_p^{init}(s)$ and $\mathcal{F}_Q^{init}(s)$ serve as a motion generator guiding the robot to acquire US images. Then the center point of the examined blood vessel $\mathcal{P}^{(i)}_{ves}$ is extracted by segmenting the blood vessel and detecting the contour where $i$ denotes the $i$-th US image. The initial center point of the examined blood vessel $\mathcal{P}^{(0)}_{ves}$ is selected manually. The nearest center point in the next US image is connected sequentially, such that the blood vessel center line $\mathcal{P}_{ves}$ is obtained. Finally, $\mathcal{P}_{ves}$ is projected to the surface of the point cloud vertically \textit{w.r.t.} $\{\mathcal{B}\}$ to get updated $\mathcal{P}$ and $\mathcal{N}$:
\begin{equation}
    \mathcal{P}^{(i)} = \arg \min_{\mathcal{P}^{(i)}\in\mathcal{P}_{arm}} \bigg|\bigg|\mathbf{N}_v( \mathcal{P}_{ves}^{(i)}-\mathcal{P}_{arm})\bigg|\bigg|^2
    \label{eq:vessel_project}
\end{equation}
where $\mathcal{P}_{arm}$ represents the point cloud of the arm phantom; $\mathcal{P}^{(i)}$ / $\mathcal{P}_{ves}^{(i)}$ denote the $i$-th sample of $\mathcal{P}$ / $\mathcal{P}_{ves}$, $\mathbf{N}_v = \mathbf{I}_{3\times3}-\mathbf{n}_{zb}\mathbf{n}_{zb}^{T}$, $\mathbf{n}_{zb}$ is the $z$-axis of $\{\mathcal{B}\}$.

\subsubsection{Path Virtual Fixture}
\par
To develop a path VF for guidance, e.g., avoiding large positional and/or rotational deviation from the optimized scanning path, a continuous trajectory ($\mathcal{F}_p(s)$, $\mathcal{F}_Q(s)$) is required. Thanks to the proposed arc-length-based path fitting algorithm, the path VF assisting clinicians freely moving the probe along the determined scan path can be easily realized by correlating the interaction force with the 1D accumulated \textit{arc-length} variable $s$.

\begin{equation}
    \begin{aligned}
        \dot{s} &= k_{pi}^{-1} \mathcal{D}zone\left(\mathcal{L}im({^\mathcal{P}f_{i}^{x}}, ~f_{\underline{i}}^{x}, ~f_{\overline{i}}^{x}),~d^x\right) \\
        s &= \mathcal{L}im\left(\int{\dot{s}} dt,~0,~1\right) \\
    \end{aligned}
    \label{eq:pHRI_position}
\end{equation}
where ${^\mathcal{P}f_{i}^{x}}={^\mathcal{P}f_{est}^{x}}-{^\mathcal{P}f_{m}^{x}}$ denotes the interaction force limited by the predefined upper and lower constraints $f_{\underline{i}}^{x}$, $f_{\overline{i}}^{x}$. ${^\mathcal{P}f_{m}^{x}}$ and ${^\mathcal{P}f_{est}^{x}}$ represent the contact forces in $\{\mathcal{P}\}$ along $x$-axis measured by the F/T sensor and estimated by the joint torque sensors, respectively. $\mathcal{D}zone(\cdot,~\cdot)$ defined by the dead-zone $d$ is introduced to suppress the disturbances such that pHRI is only activated when $|f|>d$. $k_{pi}$ determines the sensitivity to the interaction force. 

\begin{equation*}
    \begin{aligned}
        \mathcal{D}zone\left(f,~d^x\right) =&
        \begin{cases}
            0,     & if~|f|\leq d^x\\
            f-d^x, & if~f>d^x\\
            f+d^x, & if~f<-d^x\\
        \end{cases} 
    \end{aligned}
\end{equation*}
A similar interaction paradigm is applied to adjust the desired contact forces along the $z$-axis of $\{\mathcal{P}\}$ as follows:
\begin{equation}
    \begin{aligned}
        \dot{f}_d &= k_{fi}^{-1} \mathcal{D}_z\left(\mathcal{L}im({^\mathcal{P}f_{est}^{z}}-{^\mathcal{P}f_{m}^{z}}, ~f_{\underline{i}}^{z}, ~f_{\overline{i}}^{z}),~d^z\right) \\
        f_d &= \mathcal{L}im\left(\int{\dot{f}_d} dt,~f_{min},~f_{max}\right) \\
    \end{aligned}
    \label{eq:pHRI_forces}
\end{equation}

\begin{figure}[!b]
    \centering
    \includegraphics[width=0.35\textwidth, angle=0]{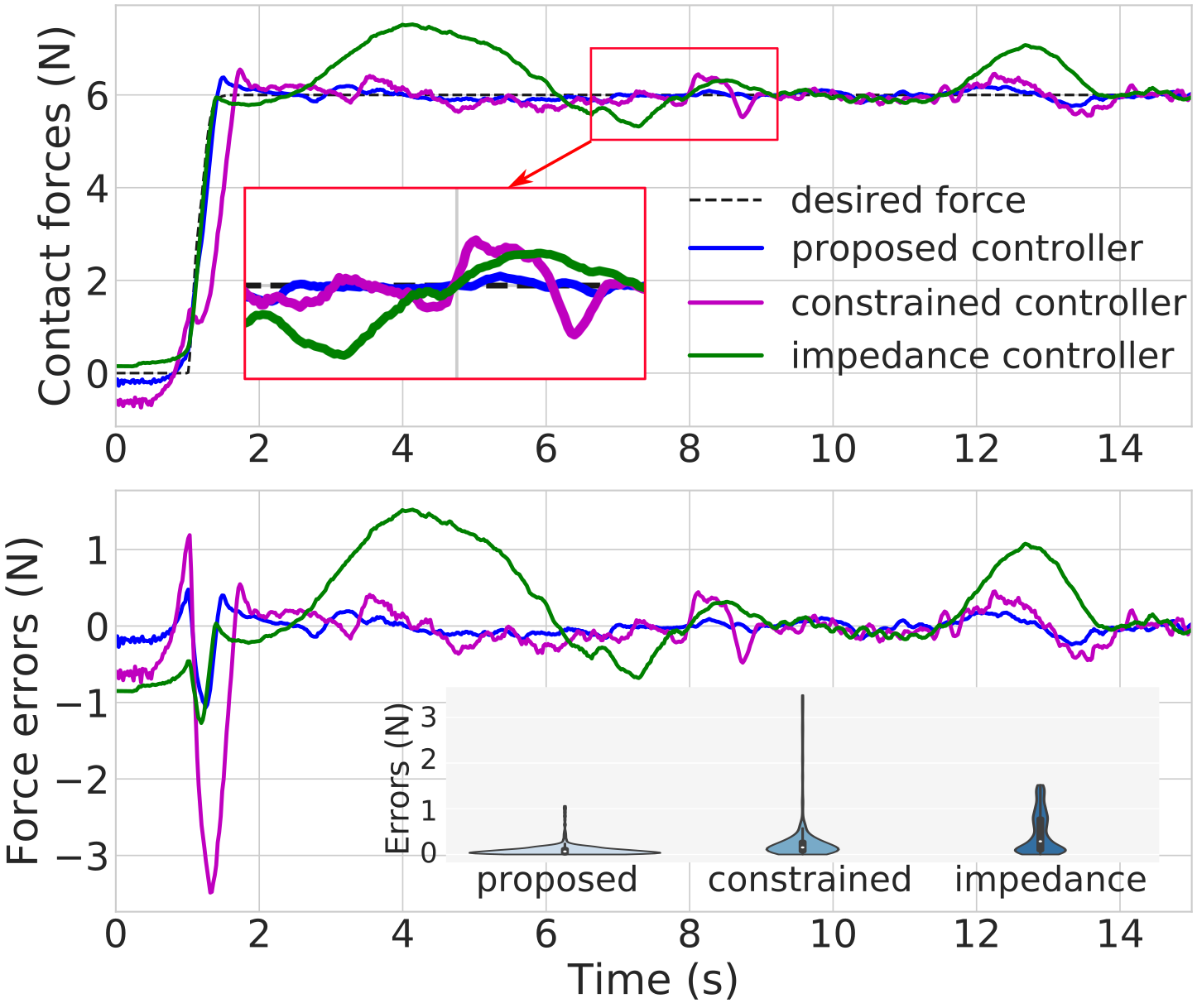}
    \vspace{-1em} 
    \caption{Comparison results of the US sweep scan performance.}
    \label{fig:sweep}
    \vspace{-1em}
\end{figure}

\begin{figure*}[!t]
     \centering
    \subfigure[sweep scan under $5 mm/s$]{
        \includegraphics[width=0.3\textwidth]{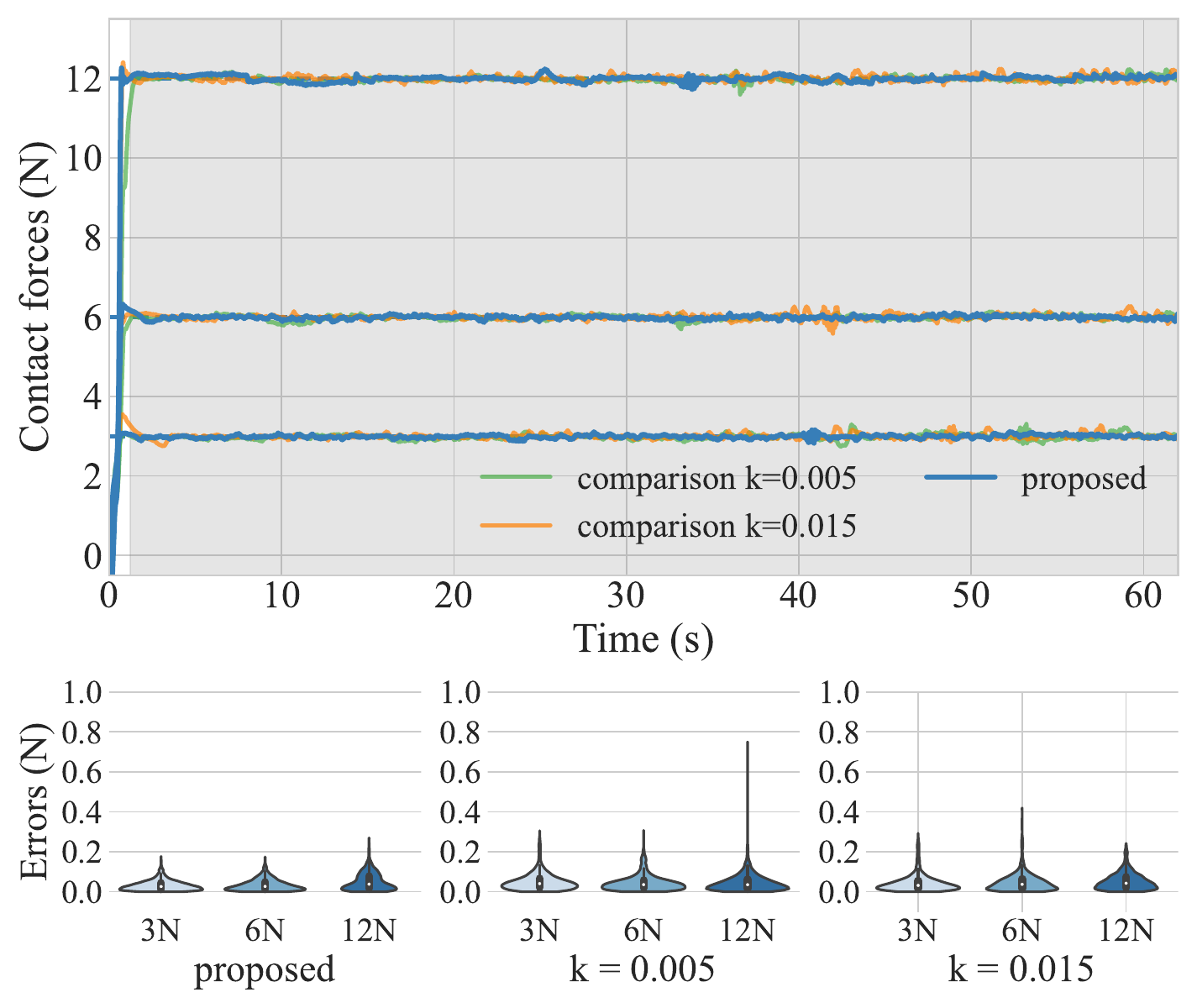}
    }
    \subfigure[sweep scan under $15 mm/s$]{
        \includegraphics[width=0.3\textwidth]{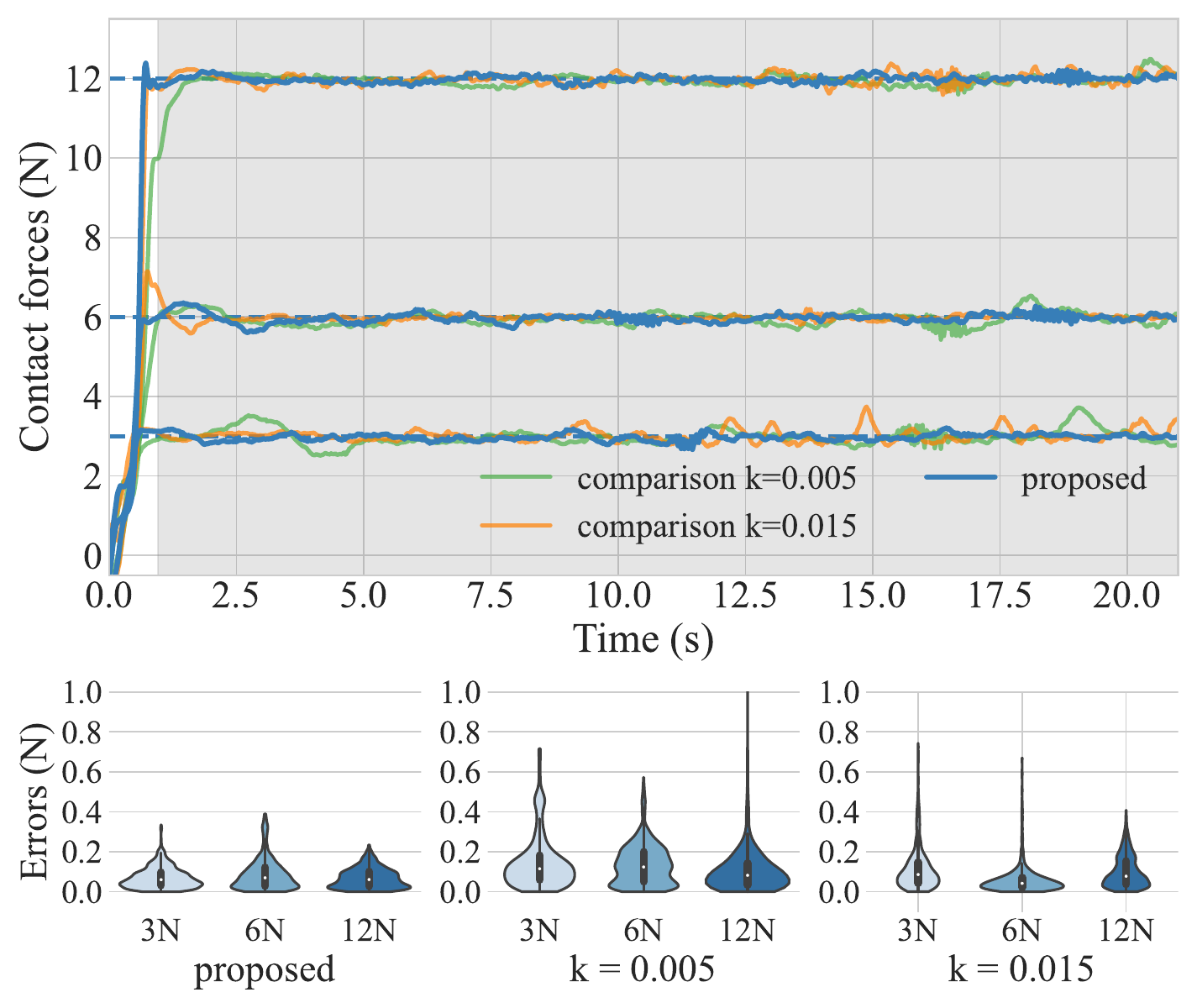}
    }
    \subfigure[sweep scan under $30 mm/s$]{
        \includegraphics[width=0.3\textwidth]{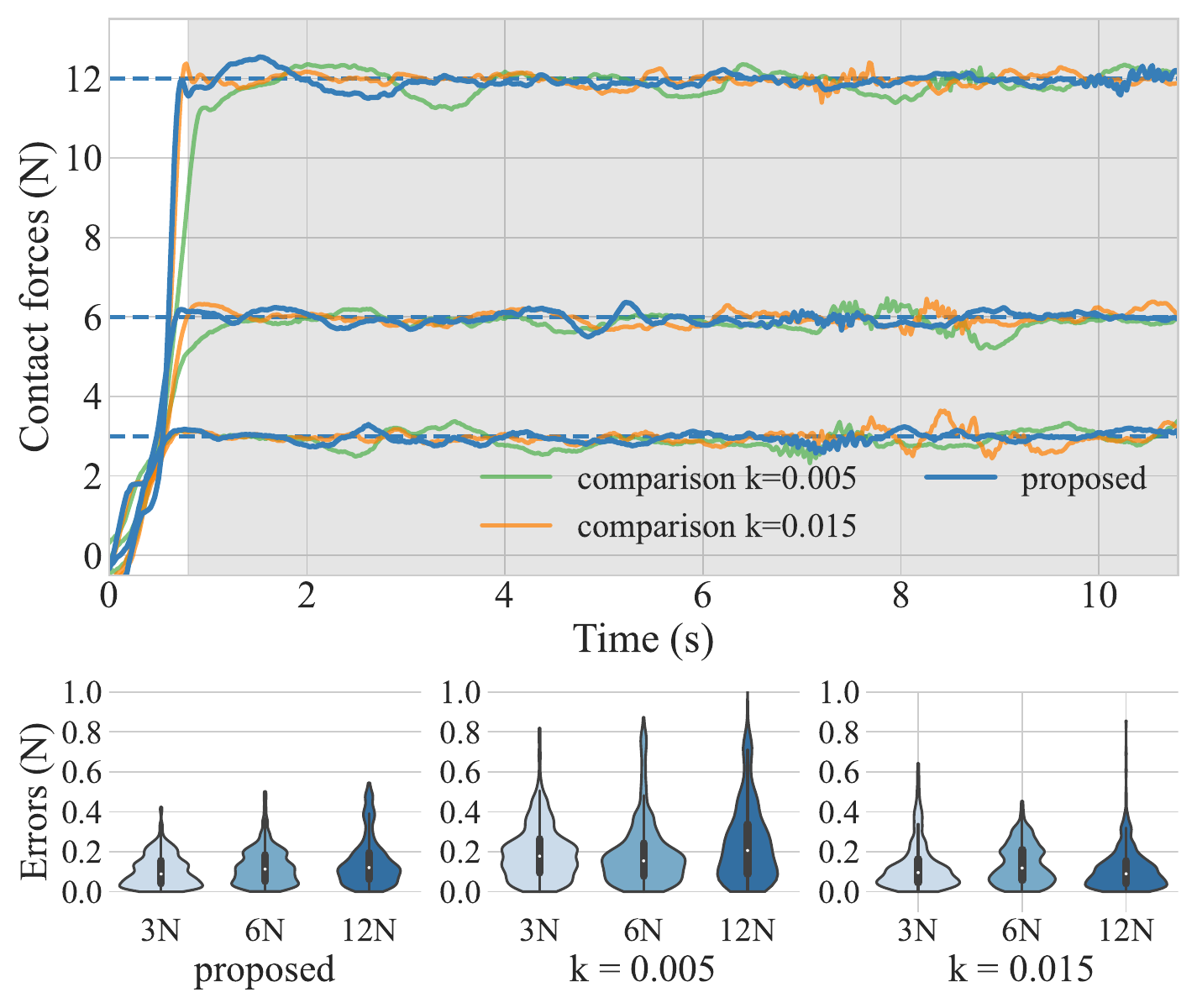}
    }
     \caption{Results for sweep scan force control performance under different scan velocities and desired contact forces. The grey shading areas indicate the time interval after contact establishment. The second row shows violin plots of absolute force control errors in the grey shading areas.}
     \label{fig:scan_ctrl_res}
\end{figure*}

\begin{figure*}[!t]
     \centering
    \subfigure[intermediate results for path planning]{
        \includegraphics[width=0.6\textwidth]{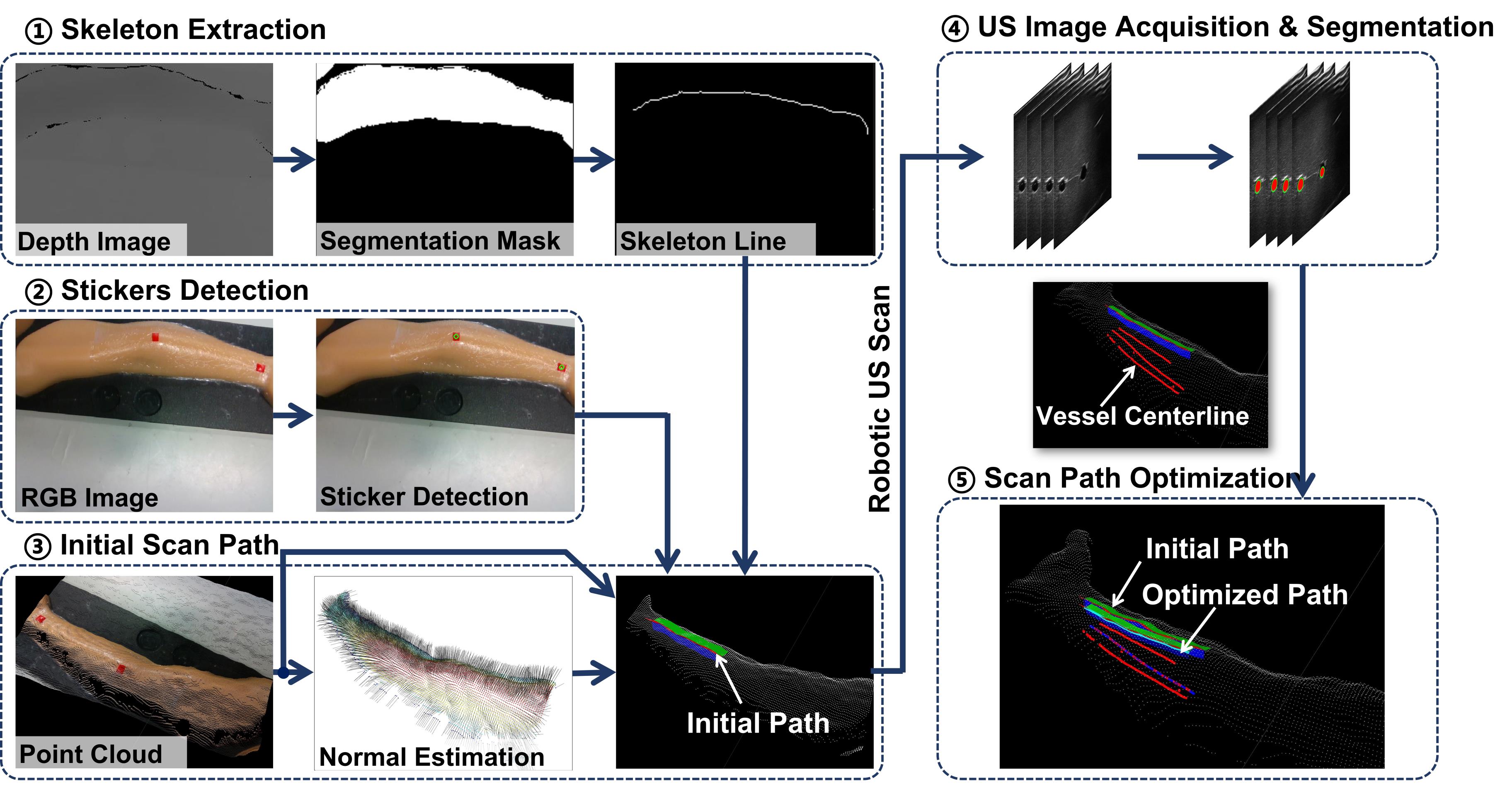}
    }
    \subfigure[center point's horizontal deviation]{
        \includegraphics[width=0.34\textwidth]{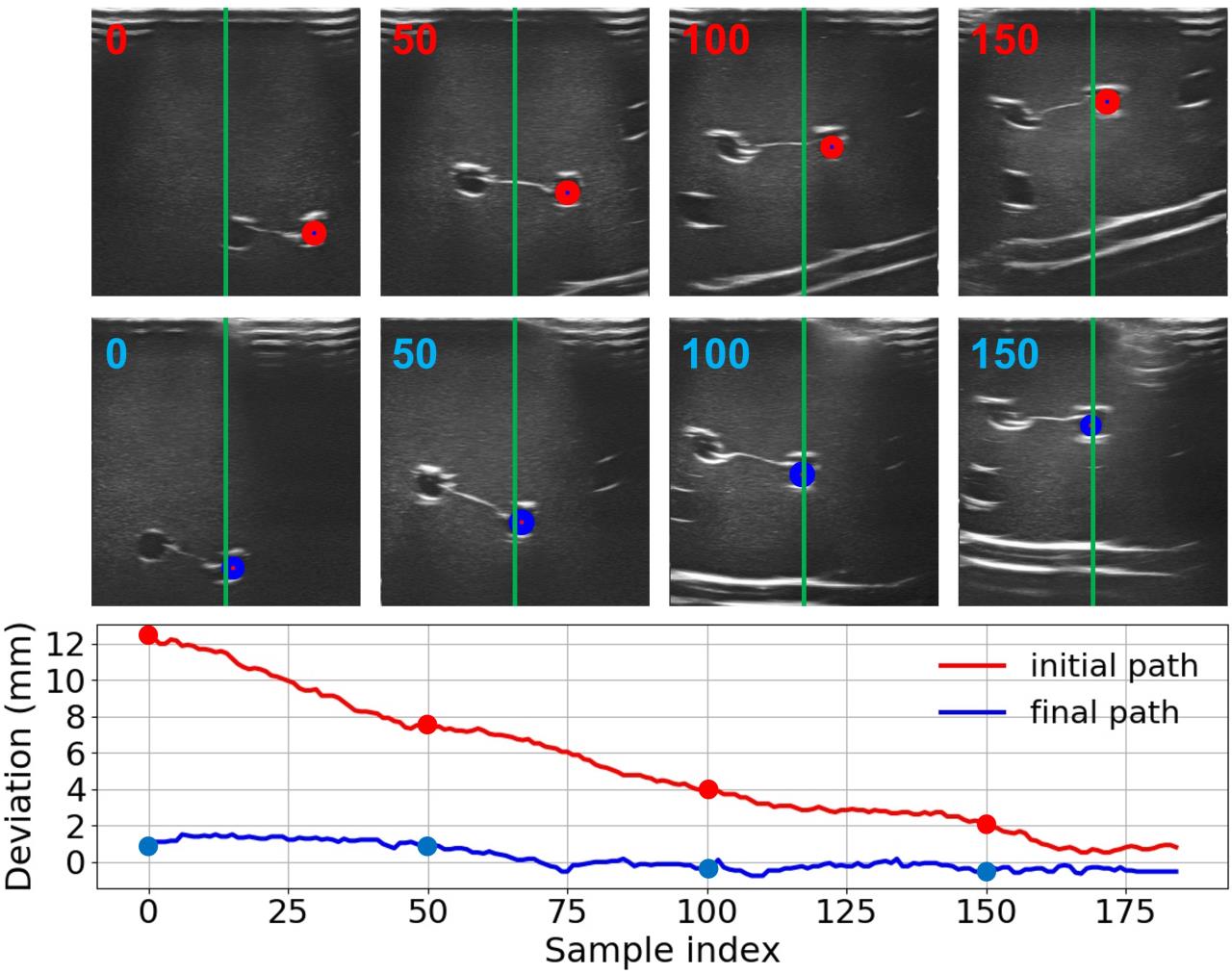}
    }
     \caption{Results for the \textit{coarse-to-fine} path planning strategy (see \textit{Section} \ref{sec:pathVF} and \textit{Alg. \ref{alg:refined_scan_path_planning}} for more details). (a) Signals flowchart and intermediate results. \textcircled{1}. The initial 2D scan path is derived by extracting the skeleton from the arm phantom mask, which is obtained through thresholding the depth image. \textcircled{2}. The starting and ending points of the scan path are identified using color stickers detected from the color image. \textcircled{3}. By estimating the normal vectors from the point cloud data and transforming the 2D pixels into 3D scan positions, an initial discrete scan path is generated. \textcircled{4}$\&$\textcircled{5}. The optimized scan path is then obtained by projecting the longitudinally distributed center points of the target vein onto the arm phantom's point cloud using Alg. \ref{alg:refined_scan_path_planning}. (b) Horizontal deviation of the examined vein's center point from the solid green line in the US image when tracking along the initial and optimized paths.}
     \label{fig:path_plan_res}
     \vspace{-1em}
\end{figure*}

\section{Experimental Results}
As shown in Fig. \ref{fig:system_setup}, the proposed RUSS for DVT exam consists of a redundant robotic manipulator (Franka Emika Panda, Franka GmbH), an RGBD camera (Intel$^\circledR$ Realsense\texttrademark ~D435), a 6-axis F/T sensor (GAMMA, SI-32-2.5/SI-65-5, ATI), a US machine (ACUSON Juniper, SIEMENS AG), a commercial arm phantom (BPA304, Blue Phantom GmbH), and a foot pedal for switching the operator's interaction intention (\emph{i.e.}, foot pedal pressed: change desired contact force; foot pedal released: change the US imaging position along the scan path). The Franka robot is controlled via the Robot Operating System (ROS) framework, running on a laptop (AMD Ryzen 9 5900HX CPU, NVIDIA GeForce RTX 3070) with Ubuntu 20.04 system. The B-mode US images are captured from the US machine via a frame grabber (MAGEWELL) at a frequency of 30 Hz. The scanning workspace of the proposed RUSS is $400~mm\times500~mm\times300~mm$ (width$\times$length$\times$height) with a maximum allowable contact force limit of $15.0N$ and a maximum automatic scanning speed set at $30~mm/s$. 

\subsection{Force Control Performance}
To demonstrate the superiority of the proposed method, the robot manipulator is commanded to perform a sweep scan on a human-like arm phantom at a velocity of $15~mm/s$ while maintaining a desired contact force of $6~N$. We compared the proposed method with the ``Impedance Controller" from \cite{dyck2022impedance} and the $log$-BLF-based ``Constrained Controller" from \cite{CASE2019ForceErr}. For the ``Impedance Controller'', the stiffness in the probe imaging direction is $100$, and the stiffness of the other dimensions is set to be the same values of the proposed method for fair comparison. The penetration offset of the scan path is $6~mm$. In the case of the ``Constrained Controller", the constraint $k_c$ is $0.4~N$ for a fair comparison. From Fig. \ref{fig:sweep}, we can observe that ``Impedance Controller" performs the worst. This result can be attributed to the varying stiffness of the phantom, as well as inaccuracies in skin surface extraction. These inaccuracies stem from the US gel covering the phantom and errors in hand-eye calibration. Consequently, the conventional ``Impedance Controller" exhibited significant tracking errors during time intervals of $3.0~s\sim6.0~s$ and $11.0~s\sim13.0~s$. In the case of ``Constrained Controller", we deliberately clip the tracking error to the range of $(-0.38,~0.38)~N$ during the implementation to ensure safety and the \textit{feasibility condition} mentioned in \cite{cao2019practical}. As a result, the ``Constrained Controller" had a longer settling time and larger tracking errors, as evident in the violin plot at the bottom right of Fig. \ref{fig:sweep}.

To further demonstrate the advantage of the proposed error transformation, we compared the force control performance of the proposed control law [Eq. (\ref{eq:ctrl_law_force})] with that of a linear feedback-only approach, denoted as $v_{f(cmp)}^\prime = k~e_f$ and referred to as the ``Fundamental Controller". Two different feedback gains are considered empirically: $k=\{0.005,~0.015\}$ with which the absolute force control errors stay within $1~N$ through out the sweep scans (see the violin plots in Fig. \ref{fig:scan_ctrl_res}]). Sweep scans are conducted on an arm phantom with an uneven surface, varying the desired contact forces $\{3,~6,~12\}~N$ and desired velocities $\{5,~15,~30\}~mm/s$.

\par
The violin plots in Fig. \ref{fig:scan_ctrl_res} demonstrate the impacts of varying desired contact force and scan velocity. It can be found that the mean and maximum force errors increase when the scan velocity increases. This observation can be explained intuitively: a faster scan velocity will result in relatively dramatic changes in contact force due to the uneven constraint surface. In addition, the results demonstrated that the changes of desired force would not significantly affect the controller's performance. We consider this is because the US gel can effectively limit the friction between probe and contact surface during scans. From Fig. \ref{fig:scan_ctrl_res} (a), we can see that the comparison controller with lower gain values $k=0.005$ has comparable performance in terms of mean errors of around $0.045~N$ at $v=5mm/s$. However, as the scan velocity increase the mean errors increase to around $0.20N$ [see Fig. \ref{fig:scan_ctrl_res} (c)]. The average settling time of the comparison controller $k=0.005$ ($0.70s$) is larger than the comparison controller $k=0.015$ ($0.40s$) and the Proposed 
Controller ($0.35s$). The transient response of the comparison controller $k=0.015$ exhibited large overshoots when $v=5mm/s$, $f_d=3~N$ and $v=15mm/s$, $f_d=6N$. 

To strike a balance between settling time and overshoot, the proposed control law [see depicted in Fig. \ref{fig:errtf}] allows users to adjust the control hyperparameters (i.e., $k_c$ and $k_s$ in Eq. (\ref{eq:err_transform})), to realize a smooth transition between the two control gains. The nonlinearity of the proposed control law achieves soft-regulation at the expense of reduced sensitivity to the disturbances within the error constraints $\pm k_c$. Although large tracking errors [see Fig. \ref{fig:scan_ctrl_res} (c), $f_d=12N$] were observed when the tracking error is less than $k_c$ ($0.4N$), these errors were effectively suppressed to less than $0.6N$ once the tracking error exceeded $\pm k_c$. Additionally, comparing the oscillations during the time periods of $11\sim16s$, $f_d=3N$ in Fig. \ref{fig:scan_ctrl_res} (b) and $7\sim9s$, $f_d=3N$ in Fig. \ref{fig:scan_ctrl_res} (c), the ``Proposed Controller" shows less oscillation and tracking error. Therefore, the proposed control law exhibits the least settling time ($0.35s$) during the contact establishment, and the maximum tracking errors remain less than $0.6N$ across all combinations of desired contact forces and scan velocities. Considering the force accuracy and response time, we consider the proposed force control suitable for US scanning on soft tissues.

\subsection{Robot-Assisted DVT-US Examination}
\label{sec:DVT_US_exam}
\begin{figure}[t]
     \centering
     \includegraphics[width=0.33\textwidth, angle=0]{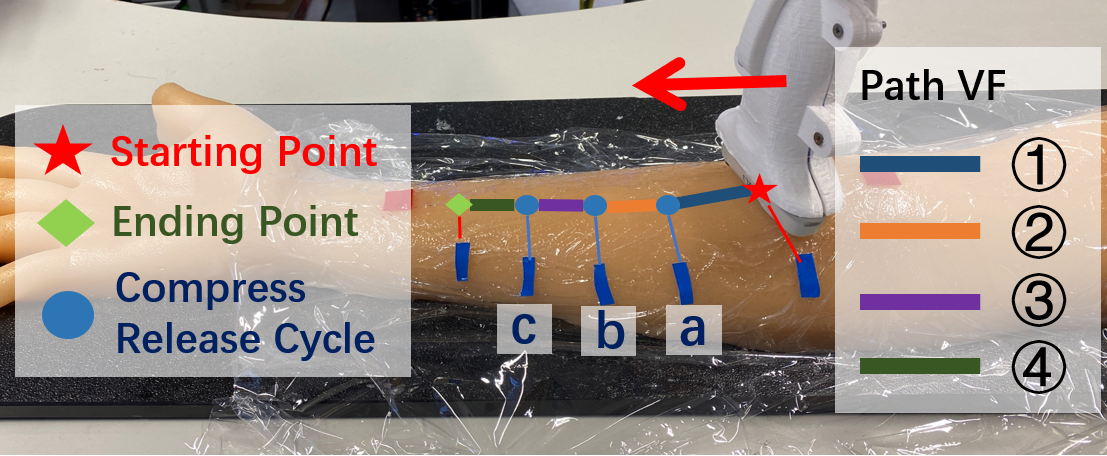}
     \vspace{-0.5em}
     \caption{An illustration of the robot-assisted DVT-US exam settings. The compression-release cycle will be executed at positions indicated by the blue points.}
     \label{fig:experiment_settings}
     \vspace{-1em}
\end{figure}

The DVT-US exam is performed using our proposed workflow. In this experiment, we evaluated the effectiveness of the derived scan path virtual fixture and the tracking performance when collaborating with clinicians. As depicted on the right-hand side of Fig. \ref{fig:overall_workflow}, once the optimized scan path is acquired, clinicians will maneuver the US probe along this path. They will pause and execute a compression-release cycle when observing a suspicious area. This procedure is repeated multiple times until the entire target vein has been thoroughly scanned. Accordingly, as presented in Fig. \ref{fig:experiment_settings}, the US probe is maneuvered along a path virtual fixture marked separately by \textcircled{1}, \textcircled{2}, \textcircled{3}, \textcircled{4} to reach three consecutive positions: $a$, $b$, and $c$, to perform the compression-release cycle.

\subsubsection{Scan Path Planning Results}

The intermediate results of the scan path planning module in a representative setup have been demonstrated in Fig.~\ref{fig:path_plan_res} (a). The coarse-to-fine path planning processes were consecutively carried out based on RGB-D images and US images, respectively. The initial and optimized paths are visualized in Fig.~\ref{fig:path_plan_res} (a). To ensure the target vessel in the US view throughout the screening, the vein's centroid was computed in real-time based on the segmented binary masks. The deviation between the computed centroid and the image's horizontal center has been calculated. In this settings, we consider that the scanning speed and contact force in a reasonable range will not result in significant difference in path accuracy. To quantitatively assess the performance variations, the deviation results obtained using initial and optimized paths are depicted in Fig.~\ref{fig:path_plan_res}~(b). 

When tracking with the initial path, notable deviations are observed (e.g., $12.5~mm$, $7.9~mm$, $4.0~mm$, $2.1~mm$ deviations at the $0$-th, $50$-th, $100$-th and $150$-th frames). This is because the location of the target vein does not align precisely with the extracted skeleton line from RGB-D images [see Fig. \ref{fig:path_plan_res}~(a)]. After the US image-based fine path adjustment, the centroid of the target vein can be much stably maintained in the horizontal middle of the US images [see the first two rows in Fig. \ref{fig:path_plan_res}~(b)]. It can be seen that the optimized path results in stable deviations ($0.8\pm0.4~mm$) across the acquisitions than the ones obtained using the initial path ($7.1\pm 3.4~mm$). Furthermore, the maximum deviation is also much smaller ($1.7~mm$ \emph{vs} $12.5~mm$). Nevertheless, the results still demonstrate that the optimized path can effectively maintain the target vein in the center of the image view, which is important for providing quantitative and repeatable diagnosis in various examinations. 

\subsubsection{Physical Human-Robot Interaction Results}
The proposed pHRI module allows clinicians to freely move the probe forward and backward along the optimized scanning path using a 6-DOF path VF. In the meantime, clinicians can perform compression exams for DVT at suspicious locations by triggering the foot pedal. The probe's motion was limited to 1-DOF (probe centerline) direction in this process. To evaluate the performance of the path VF and the tracking of varying contact forces, the validations were carried out on an arm phantom with an uneven surface. After positioning the probe on the phantom's surface, operators can only maneuver the probe along the optimized scan path with the constraint applied by the proposed path VF. The target position and force, conveyed by the operator, are updated at $50$ Hz.

\begin{figure*}[!t]
    \centering
    \includegraphics[width=0.83\textwidth]{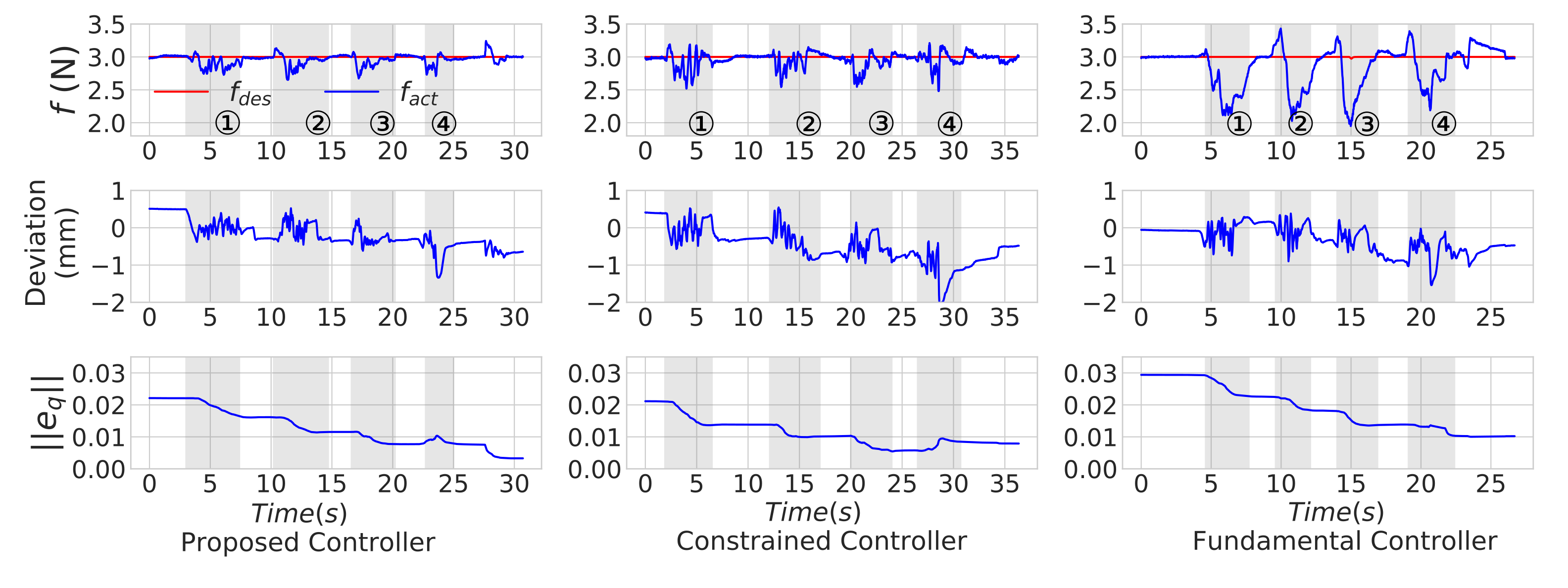}
    \vspace{-1em}
    \caption{Comparison results of the tracking performance of the VF-guided human-centric US probe motion. The grey shading areas represent the moving of the US probe.}
    \label{fig:pHRI_position}
    \vspace{-1em}
\end{figure*}

\begin{figure*}[!t]
    \centering
    \includegraphics[width=0.85\textwidth]{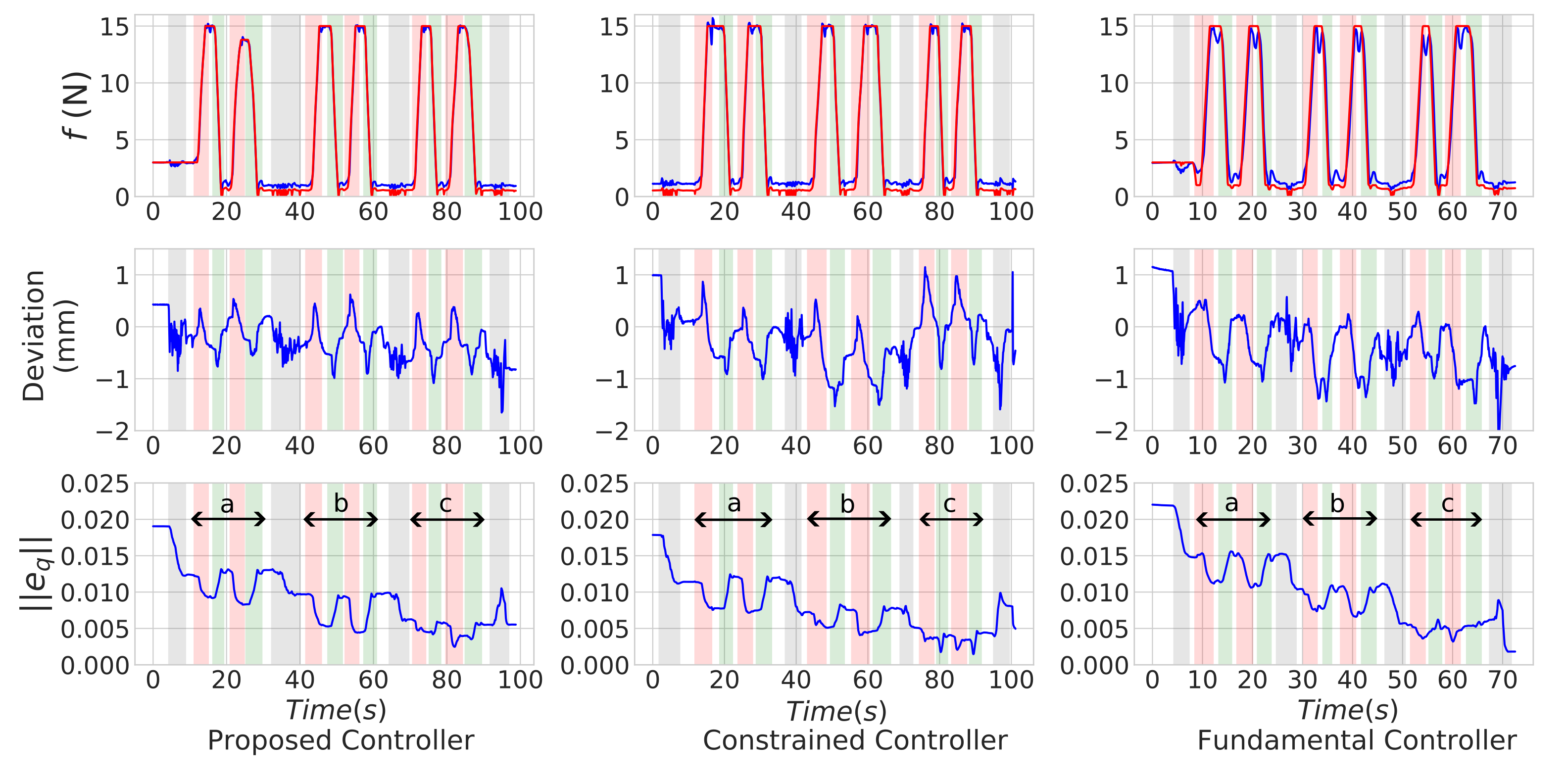}
    \vspace{-1em}
    \caption{Comparison results of the tracking performance of the VF-guided human-centric US motion for compression exam at varying imaging locations. The grey, red and green shading areas denote the moving of the probe, the compression and the release tests, respectively. (first row: red and blue lines denote desired and actual forces)}
    \label{fig:pHRI_force}
    \vspace{-1em}
\end{figure*}

\par
\begin{table}[t]
    \centering
    \caption{Statistics of force tracking results (MEAN$\pm$STD)}
    \label{tab:force_tracking}
    \vspace{-0.5em}
    \scalebox{0.95}{
    \begin{tabular}{c|c|c}
        \toprule
        \textbf{Controllers} & $\mathbf{f}_{error}$ (N) in Fig. \ref{fig:pHRI_position} & $\mathbf{f}_{error}$ (N) in Fig. \ref{fig:pHRI_force}\\
        \midrule
        Proposed & \textbf{0.099$\pm$0.079} &  \textbf{0.35$\pm$0.20} \\
        \midrule
        Constrained & 0.128$\pm$0.111 & 0.52$\pm$0.28 \\ 
        \midrule
        Fundamental & 0.435$\pm$0.269 & 1.01$\pm$0.68 \\
         \bottomrule
    \end{tabular}}
    \vspace{-1em}
\end{table}

Fig. \ref{fig:pHRI_position} presents comparison results of the tracking performance during the physical interaction under the control of the ``Proposed Controller", the ``Constrained Controller", and the ``Fundamental Controller". The clinicians continuously moved the probe along the scan path between different sections [see the Path VF in Fig. \ref{fig:experiment_settings}]. The grey shading areas correspond to the continuous movement applied by clinicians, which is constrained by the proposed path VF. Throughout the VF-guided US probe movement, the ``Proposed Controller" consistently maintains the absolute force control error less than $0.36~N$ (compared with $0.52~N$ for ``Constrained Controller" and $1.02~N$ for the ``Fundamental Controller") with the least tracking error [see Table \ref{tab:force_tracking}]. Considering the task of diagnosing DVT, clinicians will select a few locations on the scanning path to do compression examinations. The results for a representative experiment with three compression tests [see the blue dots in Fig. \ref{fig:experiment_settings}] are presented in Fig. \ref{fig:pHRI_force}. At each location, clinicians will manually repeat the compression-release cycle twice (red and green shading areas). In this context, the ``Proposed Controller" still has the least force tracking error [see Table \ref{tab:force_tracking}] and its maximum tracking error is $1.05~N$ (compared with $1.62~N$ for the ``Constrained Controller", $2.33~N$ for the ``Fundamental Controller"). The path VF tracking performance is shown in the second and third rows of Figs. \ref{fig:pHRI_position} and \ref{fig:pHRI_force}. For the compression-release cycle test in Fig. \ref{fig:pHRI_force}, we can observe that all the controllers share similar orientation tracking errors, while the proposed behaves the best in terms of tracking deviation ($0.41\pm0.27~mm$).  These findings lead to the conclusion that the proposed pHRI framework can effectively constrain the US probe to the optimized scan path during physical interaction and the proposed controller can guarantee higher force and motion tracking accuracy.

\section{Discussion}
There is limited prior research that directly addresses robotic scanning tasks or provides solutions seamlessly integrated into a robotic US system for DVT US exam. To the best of our knowledge, this is the first work that incorporates clinician protocols into the robot-assisted system for the DVT US exam. The coarse-to-fine scan path determination strategy effectively leverages external visual signals as prior information, combining them with internal US images to optimize the scan path. The proposed 6D-path fitting method eliminates the temporal information and allows users to configure arbitrary constant scan velocity on a curve scan path. Furthermore, the fitting result is a 6D-path virtual fixture over the arc-length variable $s$. These advantages collectively contribute to developing a robotic system that assists clinicians in more precisely locating the target vein and revisiting suspicious regions.

In the context of clinical use, there is a need to expand the capabilities of the robotic system to address imaging quality and vein deformation detection. A potential risk factor arises from the variability in maximum contact forces, which differs from patient to patient. To mitigate this risk, the robotic system would benefit from an integrated module capable of detecting deformation and automatically adjusting the maximum allowable forces to prevent harm to the patient. Besides, there are still some practical challenges to be addressed, such as patient's motions \cite{jiang2022towards, jiang2021motion}, and force-induced deformation \cite{abhimanyu2023unsupervised, jiang2021deformation, jiang2023defcor}. Future work of the proposed robotic system for the DVT US exam will focus on improving US image quality and enhancing patient safety.

\section{Conclusion}
This work presents a novel semi-autonomous DVT robotic US exam system. The proposed HFMC enables the accurate control of the probe to simultaneously follow both target motion and contact force commands. Besides, the proposed coarse-to-fine path planning method can stably maintain the centroid of the target vein in the horizontal middle of the US images. During the compression exam, the proposed pHRI can assist the operators in adhering the probe to the optimized scan path and performing the compression-release test at the current imaging position. The experimental results demonstrated that operators can freely move the probe along the scanning direction while the RUSS can maintain the contact force and path following performance dynamically. We believe such a semi-autonomous system with shared autonomy with clinicians is a promising solution to take both advantages of robots about the accuracy and repeatability and human operators about the advanced physiological knowledge. In the future, more tests on real patients are required to further analyze the clinical constraints and advantages.

\bibliographystyle{IEEEtran}
\bibliography{IEEEabrv, FINAL_VERSION}
\vspace{-20 mm}
\begin{IEEEbiography}[{\includegraphics[width=1in,height=1.25in,clip,keepaspectratio]{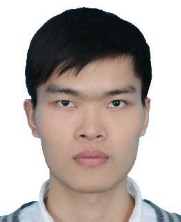}}]
{Dianye Huang} (Student Member, IEEE) received his B.Eng. degree in automation and the M.Sc. degree in control science and engineering from the School of Automation Science and Engineering, South China University of Technology, Guangzhou, China, in 2017 and 2020, respectively. He was a junior researcher with the JiHua Lab, Foshan, China. 

He is currently pursuing his doctoral degree in computer science at the Chair for Computer Aided Medical Procedures (CAMP) at the Technical University of Munich, Germany. His research interests include intelligent control, human-robot interaction, robot learning and robotic ultrasound.
\end{IEEEbiography}

\vspace{-1 mm}
\begin{IEEEbiography}[{\includegraphics[width=1in,height=1.25in,clip,keepaspectratio]{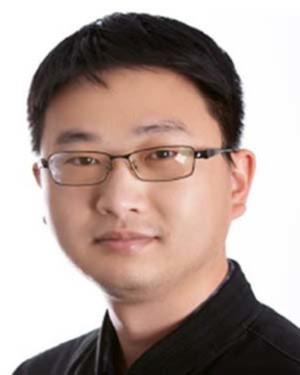}}]
{Chenguang Yang} (Senior Member, IEEE) received the Ph.D. degree in control engineering from the National University of Singapore, Singapore, in 2010. He took the postdoctoral training in human robotics from the Imperial College London, London, U.K. His research interest lies in human robot interaction and intelligent system design. 

Dr. Yang was the recipient of U.K. EPSRC UKRI Innovation Fellowship and individual EU Marie Curie International Incoming Fellowship. As the lead author, he was also the recipient of IEEE Transactions on Robotics Best Paper Award in 2012 and IEEE Transactions on Neural Networks and Learning Systems Outstanding Paper Award in 2022. He is the Co-Chair of IEEE Technical Committee on Collaborative Automation for Flexible Manufacturing and the Co-Chair of IEEE Technical Committee on Bio-mechatronics and Biorobotics Systems.
\end{IEEEbiography}

\vspace{-1 mm}
\begin{IEEEbiography}[{\includegraphics[width=1in,height=1.25in,clip,keepaspectratio]{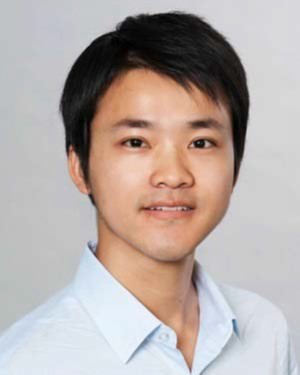}}]
{Mingchuan Zhou} (Member, IEEE) received the Ph.D. degree in computer science from the Technical University of Munich, Munich, Germany, in 2020. He was a Visiting Scholar with the Laboratory for Computational Sensing and Robotics, Johns Hopkins University, Baltimore, MD, USA, in 2019. He held a joint post-doctoral position at the Institute of Biological and Medical Imaging (IBMI), Helmholtz Center Munich, Oberschleißheim, Germany, and the Chair for Computer Aided Medical Procedures Augmented Reality (CAMP), Technical University of Munich, from 2019 to 2021. He is currently an Assistant Professor with Zhejiang University, Hangzhou, China, where he is leading multiscale robotic manipulation laboratory for agriculture. His research interests include the autonomous system, agricultural robotics, medical robotics, and image processing. 
\end{IEEEbiography}

\vspace{-10 mm}
\begin{IEEEbiography}[{\includegraphics[width=1in,height=1.25in,clip,keepaspectratio]{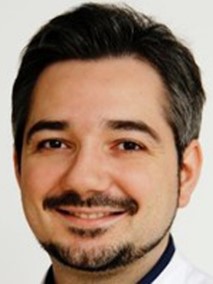}}]
{Angelos Karlas} (Member, IEEE) studied Medicine (M.D.) and Electrical and Computer Engineering (Dipl.-Ing.) at the Aristotle University of Thessaloniki, Greece. He holds a Master of Science in Medical Informatics (M.Sc.) from the same university and a Master of Research (M.Res., DIC) in Medical Robotics and Image-Guided Interventions from Imperial College London, UK. He is currently working as clinical resident at the Department for Vascular and Endovascular Surgery at the ‘rechts der Isar’ University Hospital in Munich, Germany. He is also the ‘Tenure-Track’ Group Leader of the interdisciplinary Clinical Bioengineering Group at the Helmholtz Center Munich, Germany. He completed his Ph.D. (Dr. rer. nat.) in Experimental Medicine at the Technical University of Munich, Germany. 
His main research interests are in the areas of vasometabolic and optoacoustic imaging/sensing, AI-based biomarkers as well as imageguided vascular interventions.
\end{IEEEbiography}

\vspace{-10 mm}
\begin{IEEEbiography}[{\includegraphics[width=1in,height=1.25in,clip,keepaspectratio]{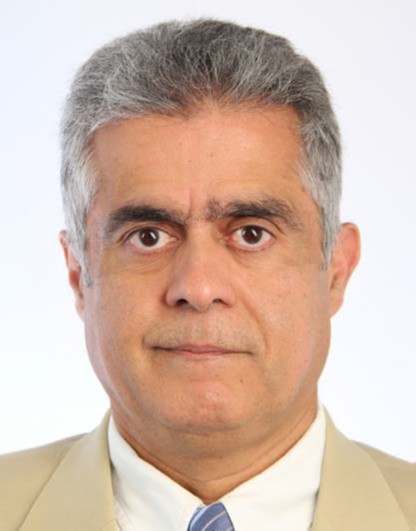}}]
{Nassir Navab} (Fellow, IEEE) received the Ph.D. degree in computer and automation from INRIA, Paris, France, and the University of Paris XI, Paris, in 1993. He is currently a Full Professor and the Director of the Laboratory for Computer-Aided Medical Procedures, Technical University of Munich, Munich, Germany, and an Adjunct Professor at Johns Hopkins University, Baltimore, MD, USA. He has also secondary faculty appointments with the both affiliated Medical Schools. He enjoyed two years of a Post-Doctoral Fellowship with the MIT Media Laboratory, Cambridge, MA, USA, before joining Siemens Corporate Research (SCR), Princeton, NJ, USA, in 1994. 
He has authored hundreds of peer-reviewed scientific articles, with more than 54 400 citations and enjoy an H-index of 104 as of August 11, 2022. He has authored more than 30 awarded papers, including 11 at the International Conference on Medical Image Computing and Computer Assisted Intervention (MICCAI), five at the International Conference on Information Processing in Computer-Assisted Interventions (IPCAI), and three at the IEEE International Symposium on Mixed and Augmented Reality (ISMAR). He is the Inventor of 50 granted U.S. patents and more than 50 International ones. 

Dr. Navab is a fellow of the Academy of Europe, MICCAI, and Asia-Pacific Artificial Intelligence Association (AAIA). He was a Distinguished Member and was a recipient of the Siemens Inventor of the Year Award in 2001 at SCR, the SMIT Society Technology Award in 2010 for the introduction of Camera Augmented Mobile C-arm and Freehand SPECT technologies, and the “10 Years Lasting Impact Award” of the IEEE ISMAR in 2015.
\end{IEEEbiography}

\vspace{-10mm}
\begin{IEEEbiography}[{\includegraphics[width=1in,height=1.25in,clip,keepaspectratio]{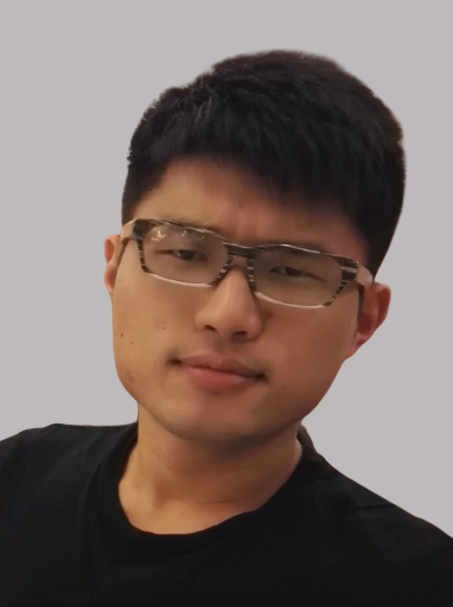}}]
{Zhongliang Jiang} (Member, IEEE) received the M.Eng. degree in Mechanical Engineering from the Harbin Institute of Technology, Shenzhen, China, in 2017, and Ph.D. degree in computer science from the Technical University of Munich, Munich, Germany, in 2022. From January 2017 to July 2018, he worked as a research assistant in the Shenzhen Institutes of Advanced Technology (SIAT) of the Chinese Academy of Science (CAS), Shenzhen, China. 

He is currently a senior research scientist at the Chair for Computer Aided Medical Procedures (CAMP) at the Technical University of Munich. His research interests include medical robotics, robot learning, human-robot interaction, and robotic ultrasound.
\end{IEEEbiography}


\end{document}